\documentclass[Afour,sageh,times]{sagej}

\usepackage{moreverb,url}

\usepackage{amsmath,amssymb,amsfonts}
\usepackage{graphicx}
\usepackage{array}
\usepackage{multirow}
\usepackage{booktabs}
\usepackage{tabularx} 
\usepackage{adjustbox} 
\usepackage{bm}
\usepackage{float}
\usepackage[table,xcdraw]{xcolor}
\usepackage{tikz}
\usetikzlibrary{positioning,shapes}

\usepackage[linesnumbered,ruled,vlined]{algorithm2e}
\usepackage{subcaption}

\usepackage{textcomp}
\usepackage{stfloats}
\usepackage{listings}
\usepackage{pdflscape} 
\usepackage{empheq}
\usepackage{todonotes}
\usepackage[final]{microtype}

\usepackage[colorlinks,bookmarksopen,bookmarksnumbered,citecolor=blue,urlcolor=blue]{hyperref}

\newcommand{\bomega}{\boldsymbol{\omega}}

\newcommand\BibTeX{{\rmfamily B\kern-.05em \textsc{i\kern-.025em b}\kern-.08em
T\kern-.1667em\lower.7ex\hbox{E}\kern-.125emX}}

\def\volumeyear{2026}

\begin{document}

\runninghead{Lang et al.}

\title{Gaussian-LIC2: LiDAR-Inertial-Camera \\ Gaussian Splatting SLAM}

\author{Xiaolei Lang$^{\dagger}$\affilnum{1}, 
Jiajun Lv$^{\dagger}$\affilnum{1}, 
Kai Tang\affilnum{1}, 
Laijian Li\affilnum{1}, 
Jianxin Huang\affilnum{1},
Lina Liu\affilnum{1}, \\
Yong Liu$^{\ast}$\affilnum{1} and
Xingxing Zuo$^{\ast}$\affilnum{2}
}

\affiliation{\affilnum{1}Institute of Cyber-Systems and Control, Zhejiang University.\\
\affilnum{2}Department of Robotics, Mohamed Bin Zayed University of Artificial Intelligence (MBZUAI).\\
$^\dag$Contributed equally. $^\ast$Corresponding authors.}

\corrauth{Yong Liu, Xingxing Zuo (Email: {\tt\small 
yongliu@iipc.zju.edu.cn,
xingxing.zuo@mbzuai.ac.ae}).}

\begin{abstract}
This paper presents a photo-realistic LiDAR-Inertial-Camera Gaussian Splatting SLAM system that simultaneously addresses visual quality, geometric accuracy, and real-time performance. The proposed method performs robust and accurate pose estimation within a continuous-time trajectory optimization framework, while incrementally reconstructing a large-scale 3D Gaussian map using camera and LiDAR data, all in real time. The resulting map enables high-quality, real-time novel view rendering of both RGB images and depth maps.
To effectively address under-reconstruction in regions not covered by the LiDAR, we employ a lightweight zero-shot depth model that synergistically combines RGB appearance cues with sparse
LiDAR measurements to generate dense depth maps. The depth completion enables reliable Gaussian initialization in LiDAR-blind areas, significantly improving system applicability for sparse LiDAR sensors.
To enhance geometric accuracy, we fully use sparse but precise LiDAR depths to supervise Gaussian map optimization and accelerate it with carefully designed CUDA-accelerated strategies. Furthermore, we explore how the incrementally reconstructed Gaussian map can improve the robustness of the odometry. By efficiently incorporating photometric constraints from the Gaussian map into the continuous-time factor graph optimization, we demonstrate improved tracking performance, especially under LiDAR degenerated scenarios.
We also showcase downstream applications via extending our elaborate system, including video frame interpolation and fast 3D mesh production.
To support rigorous evaluation, we collect a dedicated LiDAR-Inertial-Camera dataset featuring ground-truth poses, depth maps, and extrapolated trajectories for assessing out-of-sequence novel view synthesis in large-scale outdoor scenarios. Extensive experiments on both public and self-collected datasets demonstrate the superiority and versatility of our system across LiDAR sensors with varying sampling densities. Both the dataset and code will be made publicly available on the project page \url{https://xingxingzuo.github.io/gaussian_lic2}.

\end{abstract}


\maketitle

\section{Introduction}

\begin{figure}[t!]
    \centering
    \includegraphics[width=\linewidth]{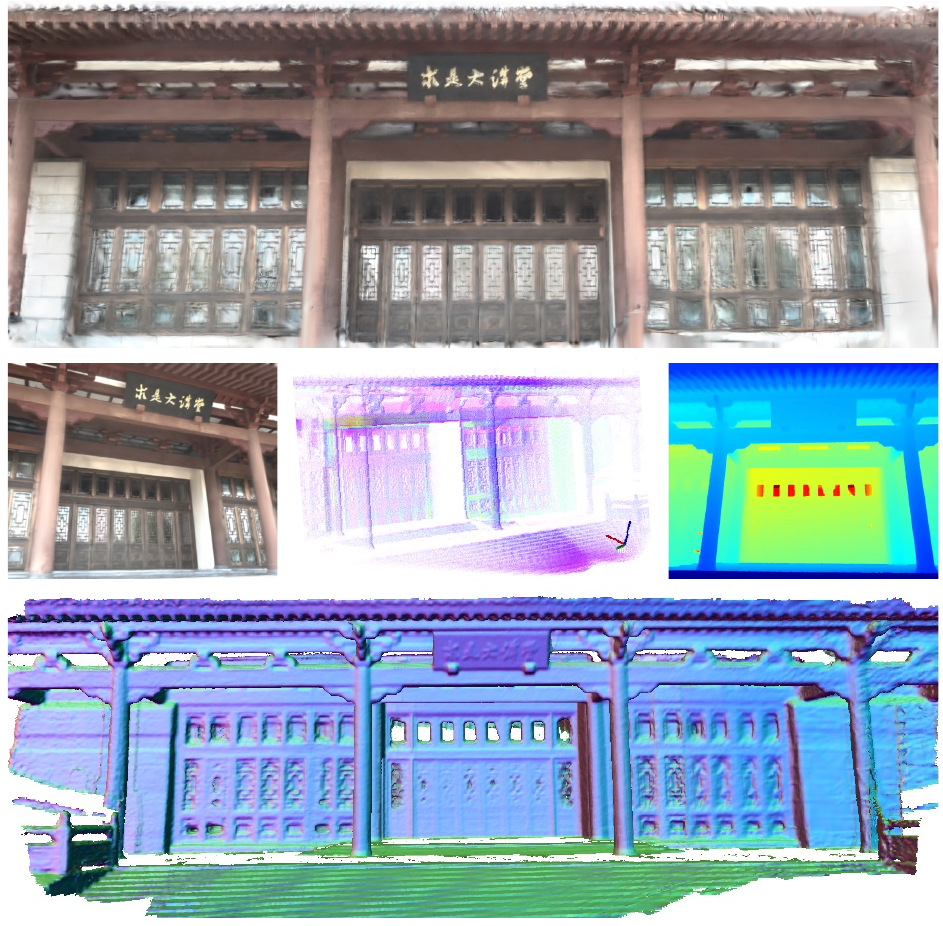}
    \caption{
    Overview of the system outputs (ordered left to right, top to bottom): (1) reconstructed Gaussian map, (2) RGB image rendered from a novel viewpoint, (3) sparse LiDAR point cloud map, (4) depth map rendered from the novel viewpoint, and (5) 3D mesh extracted from the Gaussian map.
    }
    \label{fig:teaser}
\end{figure}

Simultaneous localization and mapping (SLAM) serves as a fundamental technology that facilitates spatial perception in both mixed reality systems and robotic applications. Recent advances in radiance field representations, particularly Neural Radiance Fields (NeRF)~\cite{mildenhall2021nerf} and 3D Gaussian Splatting (3DGS)~\cite{kerbl20233d}, have pioneered a new paradigm in SLAM, namely radiance-field-based SLAM~\cite{tosi2024nerfs}. Powered by differentiable photo-realistic rendering, radiance field-based SLAM systems aim to provide both accurate pose estimation and photo-realistic 3D maps in real time, equipping robots with richer 3D scene understanding. These systems can benefit a wide range of tasks, including path planning~\cite{jin2024gs, chen2025splat, lei2025gaussnav}, active exploration and mapping~\cite{chen2025activegamer}, and 3D mesh reconstruction~\cite{zhang2024rapid}. 

Initially, NeRF-based SLAM systems~\cite{sucar2021imap, zhu2022nice, yang2022vox, wang2023co, johari2023eslam} leverage multi-layer perceptrons (MLPs), often further integrating learnable encodings, to represent the entire scene and generate high-quality dense maps with low memory consumption.
Nonetheless, such implicit representations necessitate computationally intensive volumetric rendering based on sampling in 3D space,
undermining the real-time capability, which is essential for robot applications. The emergence of 3DGS has shifted this landscape. It features fast rendering along with {\color{blue} good visual quality} and demonstrates greater potential for real-time use. Equipped with RGB-D or RGB sensors, most 3DGS-based SLAM systems~\cite{yan2024gs, keetha2024splatam, yugay2023gaussian, matsuki2024gaussian, huang2024photo} focus on indoor environments and outperform NeRF-based approaches, but unfortunately struggle in challenging conditions such as violent ego-motion, varying illumination, and lack of
visual textures, commonly arising in unbounded outdoor scenes.  Several works~\cite{sun2024mm3dgs, wu2024mm, xiao2024liv, xie2024gs, zhao2025lvi, hong2025gs} and our previous work~\cite{lang2025gaussian} address the issue by fusing LiDAR and IMU, among which the effectiveness of LiDAR-Inertial-Camera fusion has been extensively validated in traditional SLAM methods~\cite{zhang2018laser, zuo2019lic, zuo2020lic, shan2021lvi, lin2022r, lin2024r, zheng2022fast, zheng2024fast, yuan2024sr, lv2023continuous, lang2023coco, cong2024or, wu2025dali}. Notably, the precise geometric priors provided by LiDAR contribute to both pose tracking and Gaussian mapping, and the significantly reduced cost of LiDARs nowadays has made them much more accessible for integration.

Despite recent efforts to incorporate Gaussian Splatting~\cite{kerbl20233d} into LiDAR-Inertial-Camera SLAM systems~\cite{lang2025gaussian,xiao2024liv,xie2024gs,zhao2025lvi,hong2025gs}, several key issues remain unsolved or insufficiently addressed. \textbf{First}, current methods predominantly rely on the dense LiDAR with the FoV closely aligned with that of the camera, initializing Gaussians solely from LiDAR points. This can lead to under-reconstruction in LiDAR blind spots, especially when the sparse LiDAR is used.
Although adaptive density control (ADC) can alleviate this limitation by cloning and splitting Gaussians based on the magnitude of back-propagated gradients and scales of Gaussians ~\cite{kerbl20233d}, it struggles in the incremental SLAM systems involving optimization of Gaussians with different maturity. ADC also requires aggregating gradients over multiple iterations of optimization, which leads to a lack of timeliness for real-time SLAM systems. 
\textbf{Second}, despite leveraging LiDAR data with precise geometric information, existing SLAM methods overemphasize the visual quality of the map while neglecting its geometric reconstruction quality, which limits its applicability in geometry-critical tasks, like obstacle avoidance. Pursuing both high-quality RGB and depth rendering while maintaining real-time performance remains a great challenge. \textbf{Third}, current approaches usually decouple LiDAR-Inertial-Camera odometry and Gaussian mapping. It is underexplored whether the incrementally constructed Gaussian map can benefit the traditional odometry during system operation. 
\textbf{Furthermore}, most methods focus merely on rendering quality from training views, overlooking the novel view synthesis capability in sequence or even out of sequence. Admittedly, LiDAR-Inertial-Camera SLAM datasets that enable joint evaluation of RGB and depth rendering across both in-sequence and out-of-sequence novel views are still scarce.

Evolved from our conference version, Gaussian-LIC~\cite{lang2025gaussian}, this journal extension addresses the aforementioned challenges by proposing a real-time, geometry-aware, and photo-realistic SLAM system, termed Gaussian-LIC2. The proposed system achieves robust and accurate pose estimation while constructing photorealistic 3D Gaussian maps that capture both high-fidelity visual appearance and precise geometric structure. Main contributions are as follows:
\begin{itemize}
	\item We propose the first LiDAR-Inertial-Camera Gaussian Splatting SLAM system that jointly takes care of visual quality, geometric accuracy, and real-time performance. It is capable of robustly and precisely estimating poses while constructing a photo-realistic and geometrically accurate 3D Gaussian map, all in real time.
	\item 
    We propose integrating LiDAR and visual data through an efficient, lightweight, generalizable sparse depth completion network to rapidly predict depths for image patches not covered by the LiDAR, enabling more comprehensive 3D Gaussian initialization and mitigating under-reconstruction. During training, we fully leverage LiDAR-provided depth for supervision and accelerate the process with a series of carefully designed C++ and CUDA implementations.
    \item 
    We explore jointly fusing photometric constraints from the incrementally built Gaussian map with raw LiDAR-Inertial data in a continuous-time framework, successfully helping the odometry overcome the LiDAR degeneration. Besides, we have extended our system to enable Gaussian map utilization for downstream applications such as video frame interpolation and rapid mesh generation.
	\item We curate a specialized LiDAR-Inertial-Camera dataset with ground-truth poses and depths, as well as carefully designed capturing trajectories, enabling the evaluation of out-of-sequence novel view synthesis. We conduct extensive experiments on public and self-collected datasets, proving the superiority of our approach and its adaptability to various types of LiDARs. 
\end{itemize}

\section{Related Works}
\label{sec:related_work}

\subsection{Photo-Realistic Reconstruction with Radiance Field}
Map representation deeply affects architecture designs and potential downstream applications of SLAM systems. Sparse SLAM approaches~\cite{qin2018vins, campos2021orb} excel in pose estimation but yield only sparse keypoint maps. On the contrary, dense SLAM methods produce dense maps beneficial for scene understanding. For instance, DTAM~\cite{newcombe2011dtam}, REMODE~\cite{pizzoli2014remode}, LSD-SLAM~\cite{engel2014lsd}, DSO~\cite{engel2017direct}, DROID-SLAM~\cite{teed2021droid}, etc, can achieve accurate camera pose tracking and reconstruct dense point cloud maps.
KinectFusion~\cite{newcombe2011kinectfusion}, ElasticFusion~\cite{whelan2015elasticfusion}, and SurfelMeshing~\cite{schops2019surfelmeshing} model 3D scenes using truncated signed distance function (TSDF), surfel, and mesh, all of which are commonly utilized in the SLAM field. However, it is challenging to recover photo-realistic camera views from these map representations. Fortunately, the emergence of NeRF~\cite{mildenhall2021nerf} has brought a promising solution. As a novel radiance field representation, NeRF implicitly models the geometry and texture of the scene through MLPs and achieves differentiable photo-realistic rendering even at novel views. 
To accelerate the training, Plenoxels~\cite{fridovich2022plenoxels} optimizes spherical harmonics and density directly on a sparse voxel grid without neural networks, and instant-ngp~\cite{muller2022instant} introduces a multiresolution hash encoding paired with a tiny GPU-accelerated neural network.
To improve the geometric accuracy of NeRF representation and extract a precise 3D mesh from it, VolSDF~\cite{yariv2021volume}, NeuS~\cite{wang2021neus}, and NeuS2~\cite{wang2023neus2} replace the density field with signed distance function (SDF). 

However, NeRF still falls slightly short in real-time reconstruction due to the computationally intensive ray-based volume rendering. In contrast, 3D Gaussian Splatting (3DGS)~\cite{kerbl20233d} explicitly represents scenes using view-dependent anisotropic Gaussians and introduces a tile-based rasterization strategy for splats, achieving faster rendering speed while maintaining superior visual quality. Built upon 3DGS, a number of follow-up works have been proposed to further enhance its performance and flexibility. Scaffold-GS~\cite{lu2024scaffold} incorporates tiny MLPs to predict the properties of neural Gaussians, enabling more compact and expressive representations. Meanwhile, SuGaR~\cite{guedon2024sugar} and GOF~\cite{yu2024gaussian} focus on efficient mesh extraction from 3DGS representations, facilitating downstream tasks such as geometry processing and simulation. 2DGS~\cite{huang20242d}, GaussianSurfels~\cite{dai2024high}, and PGSR~\cite{chen2024pgsr} flatten the Gaussians to accurately conform to the scene surface, while RaDe-GS~\cite{zhang2024rade} introduces an enhanced depth rasterizing approach without the reformulation of the Gaussian primitives. NeuSG~\cite{chen2023neusg} and GSDF~\cite{yu2024gsdf} try to combine neural SDF with 3DGS using mutual guidance and joint supervision.

Original NeRF and 3DGS methods typically rely solely on image data for photo-realistic reconstruction. However, the incorporation of LiDAR, which has become increasingly affordable and accessible, can significantly enhance the performance of both NeRF and 3DGS, particularly in unbounded outdoor environments.
For NeRF, LiDAR data can guide ray sampling around surfaces and provide accurate depth supervision during optimization. 
URF~\cite{rematas2022urban} and EmerNeRF~\cite{yang2023emernerf} are NeRF variants tailored for autonomous driving scenarios and achieve outstanding rendering quality with the aid of LiDAR. Bootstrapped by a LiDAR SLAM system, SiLVR~\cite{tao2024silvr} constructs multiple NeRF submaps efficiently. M2Mapping~\cite{liu2024neural} further unifies surface reconstruction and photo-realistic rendering in LiDAR-Camera systems through an SDF-based NeRF formulation.
In the context of 3DGS, LiDAR data can be leveraged not only for geometric supervision but also for the accurate and efficient initialization of Gaussians. PVG~\cite{chen2023periodic}, DrivingGaussian~\cite{zhou2024drivinggaussian}, StreetGaussians~\cite{yan2024street}, and TCLC-GS~\cite{zhao2024tclc} are the first to introduce LiDAR into the 3DGS framework, effectively modeling both static scenes and dynamic objects in autonomous driving environments. LIV-GaussMap~\cite{hong2024liv} and LetsGo~\cite{cui2024letsgo} both utilize the point cloud and poses from LiDAR-based SLAM to initialize 3DGS. Focusing more on geometric quality, LI-GS~\cite{jiang2024li} employs 2DGS as the map representation to enhance surface alignment, while GS-SDF~\cite{liu2025gs} incorporates LiDAR into SDF-based 3DGS.

The aforementioned works are all per-scene optimization frameworks. Interestingly, a series of feed-forward models~\cite{yu2021pixelnerf, chen2021mvsnerf, charatan2024pixelsplat, chen2024mvsplat} have emerged and achieved generalizable photo-realistic reconstruction in an end-to-end manner. However, their accuracy still lag behind that of per-scene optimization methods, and their applicability is limited to a small number of high-quality images with minimal viewpoint variation.
\vspace{-1em}

\subsection{Incremental Visual SLAM Systems with Radiance Field}
Radiance-field-based reconstruction, whether per-scene optimized or feed-forward, is essentially an offline process with all the collected data accessible. In comparison, radiance-field-based SLAM systems incrementally perform photo-realistic reconstruction with sequentially input sensor data, utilizing radiance file map representations. 

Given sequential RGB-D inputs, iMAP~\cite{sucar2021imap} is a pioneering work built upon the implicit neural representation to achieve watertight online reconstruction. Following this, NICE-SLAM~\cite{zhu2022nice} is able to scale up to larger indoor scenes by combining MLP representation with hierarchical feature grids. Vox-Fusion~\cite{yang2022vox} further adopts the octree to dynamically expand the volumetric map, eliminating the need for pre-allocated grids. Leveraging hash grids, tri-planes, and neural point cloud as their respective implicit representations, Co-SLAM~\cite{wang2023co}, ESLAM~\cite{johari2023eslam}, and Point-SLAM~\cite{sandstrom2023point} get enhancement in both localization and mapping. 
Moreover, H2-Mapping~\cite{jiang2023h} handles the forgetting issue by a novel coverage-maximizing keyframe selection strategy.
In terms of 3DGS-based SLAM, GS-SLAM~\cite{yan2024gs}, SplaTAM~\cite{keetha2024splatam}, and Gaussian-SLAM~\cite{yugay2023gaussian} demonstrate the advantages of 3DGS
over existing map representations in SLAM systems for online photo-realistic mapping. 
By forcing binary opacity for each Gaussian, RTG-SLAM~\cite{peng2024rtg} achieves real-time performance indoors with the compact scene representation. GSFusion~\cite{wei2024gsfusion} jointly constructs the Gaussian map with a TSDF map and uses a quadtree data structure to reduce the number of Gaussians. MM3DGS-SLAM~\cite{sun2024mm3dgs} loosely fuses RGB-D and IMU measurements to enable more robust and precise pose estimation.

A range of studies have also explored operating solely on monocular RGB images, among which NICER-SLAM~\cite{zhu2024nicer} and MonoGS~\cite{matsuki2024gaussian} are the representative radiance-field-map-centric approaches. The former fully makes use of the monocular geometric cues for supervision, and the latter introduces the isotropic regularization to address ambiguities in incremental reconstruction. However, decoupled tracking and mapping methods, which adopt the state-of-the-art visual odometry for pose tracking and radiance field optimization for photo-realistic mapping in parallel, usually demonstrate much more robustness. Orbeez-SLAM~\cite{chung2023orbeez} and Photo-SLAM~\cite{huang2024photo} are both built upon ORB-SLAM~\cite{campos2021orb}. NeRF-SLAM~\cite{rosinol2023nerf} and IG-SLAM~\cite{sarikamis2024ig} utilize the dense depth maps estimated from the tracking front-end DROID-SLAM as additional information to supervise the training of instant-ngp and 3DGS. NeRF-VO~\cite{naumann2024nerf} and MGS-SLAM~\cite{zhu2024mgs} instead employ the sparse DPVO~\cite{teed2023deep} as a faster tracker with network-predicted dense depth maps for supervision.

\subsection{LiDAR-Integrated SLAM Systems with Radiance Field}
Radiance-field-based Visual SLAM has achieved great performance in confined indoor scenes, and fusing IMU data can further improve the robustness. But they are still challenged by extreme violent motions, drastic lighting changes, and texture deficiency. Studies in both radiance-field-based reconstruction~\cite{rematas2022urban, yang2023emernerf, tao2024silvr, liu2024neural, chen2023periodic, zhou2024drivinggaussian, yan2024street, zhao2024tclc, hong2024liv, cui2024letsgo, jiang2024li, liu2025gs} and conventional SLAM~\cite{zhang2018laser, zuo2020lic, shan2021lvi, lin2022r, lin2024r, zheng2022fast, zheng2024fast, yuan2024sr, lv2023continuous, lang2023coco} have validated the superiority of introducing LiDARs. For radiance-field-based SLAM, the advantages of integrating LiDAR include, at the very least, enhanced robustness and accuracy in pose estimation, precise geometric supervision, facilitating efficient sampling for NeRF and accurate initialization for 3DGS.

Powered by the LiDAR data, SHINE-Mapping~\cite{zhong2023shine} is an incremental mapping framework that uses octree-based hierarchical neural SDF to perform large-scale 3D reconstruction in a memory-efficient way. Meanwhile, NF-Atlas~\cite{yu2023nf} organizes multiple neural submaps by a pose graph, and N3-Mapping~\cite{song2024n} applies a voxel-oriented sliding window mechanism to alleviate the forgetting issue with a bounded memory footprint. Instead of purely geometric mapping, HGS-Mapping~\cite{wu2024hgs} incrementally builds a dense photo-realistic map with hybrid Gaussians in urban scenarios. All these mapping systems assume a priori ground-truth poses, whereas full SLAM systems that estimate poses simultaneously are inherently more complicated. NeRF-LOAM~\cite{deng2023nerf} is a typical NeRF-based LiDAR odometry and mapping method, optimizing poses and voxel embeddings concurrently. Furthermore, LONER~\cite{isaacson2023loner} proposes a novel information-theoretic loss function to attain real-time performance. Benefiting from a semi-explicit representation, PIN-SLAM~\cite{pan2024pin} utilizes a neural point cloud map representation with elasticity for globally consistent mapping. Splat-LOAM~\cite{giacomini2025splat} exploits explicit 2DGS primitives with spherical projection for localization and LiDAR mapping.

Going beyond merely recovering the geometric structure of the scene, Rapid-Mapping~\cite{zhang2024rapid} utilizes NeRF to represent and render detailed scene textures. MM-Gaussian~\cite{wu2024mm}, LiV-GS~\cite{xiao2024liv}, and PINGS~\cite{pan2025pings} perform coupled pose tracking and dense mapping within a unified 3DGS optimization framework using LiDAR-Camera data. While these methods demonstrate impressive performance, they are not capable of real-time operation, despite the inherent efficiency of 3DGS. By comparison, decoupled tracking and Gaussian mapping approaches have shown promising results. Our prior work Gaussian-LIC~\cite{lang2025gaussian} is the first real-time LiDAR-Inertial-Camera 3DGS-based SLAM system, combining continuous-time LiDAR-Inertial-Camera odometry for tracking with 3DGS for mapping, achieving high-fidelity photo-realistic reconstruction in real time. Subsequently, GS-LIVM~\cite{xie2024gs}, LVI-GS~\cite{zhao2025lvi}, and GS-LIVO~\cite{hong2025gs} have also been built atop classical LiDAR-Inertial-Camera odometry.  Among them, GS-LIVM and GS-LIVO both adopt voxel-based map structures and initialize Gaussian covariances.  

\subsection{Extension over Gaussian-LIC}
This paper is evolved from our conference version, Gaussian-LIC~\cite{lang2025gaussian}, and further introduces important systemic and technical contributions beyond it.

\textbf{Systemically}, to the best of our knowledge, Gaussian-LIC2 is the first LiDAR-Inertial-Camera Gaussian Splatting SLAM system which harmonizes high-fidelity visual realism, accurate geometry, and real-time efficiency. In contrast, Gaussian-LIC, as well as GS-LIVM, and GS-LIVO, focus primarily on visual quality or real-time performance at most.

\textbf{Technically}, 
there are enhancements in three key aspects:
\begin{itemize} 
\item \textit{Accommodate LiDAR Blind Area}: 
GS-LIVM, LVI-GS, and GS-LIVO create Gaussians solely from LiDAR data, which can lead to under-reconstruction in LiDAR blind areas. Gaussian-LIC alleviates this limitation by incorporating visual features triangulated online; however, it struggles in textureless scenes and is sensitive to parallax between views, which degrades triangulation reliability. Gaussian-LIC2 leverages visual cues more effectively by fusing RGB appearance with sparse yet geometrically accurate LiDAR measurements through an efficient depth completion model. This enables the generation of accurate dense depth maps and helps better initialize Gaussians for carefully selected pixels unobserved by the LiDAR. Further insights are provided in Sec.~\ref{sec:dc}. 
\item \textit{Geometric Accuracy with Visual Quality and Real-Time Performance as Prerequisites}: 
Visual fidelity, real-time efficiency, and geometric accuracy — satisfying any single one is readily achievable, but balancing all three is challenging. As a LiDAR-Inertial-Camera 3DGS-based SLAM system, Gaussian-LIC is the first to attain high visual quality in real time. Nevertheless, geometric accuracy, which affects the quality of rendered depth and the robustness of novel view synthesis, has been neglected by Gaussian-LIC and other approaches. Some attempts, such as flattened primitives or extra regularization terms, can enhance the geometry but tend to hinder the visual quality or real-time performance. To this end, Gaussian-LIC2 advocates achieving the most effective regularization under minimal supervision. It maximally exploits sparse LiDAR depth for map regularization and further enhances CUDA optimization strategies in Gaussian-LIC, enabling faster forward and backward propagation of depth rasterization. More importantly, the incremental
insertion of Gaussians in SLAM can lead to depth underestimation for certain pixels. Gaussian-LIC2 mitigates this issue by scaling the rendered depth according to the rendered opacity, ensuring consistent and accurate depth supervision.

\item \textit{Gaussian Map Feedback to Conventional Odometry}: 
The odometry and Gaussian mapping modules are fully decoupled in Gaussian-LIC and many existing methods. Gaussian-LIC2 investigates whether the incrementally built Gaussian map can, in turn, enhance the odometry performance by introducing a photometric constraint derived from the Gaussian map. Notably, Gaussian-LIC2 alleviates the impact of the varying camera exposure on the photometric constraint and validates the critical role of geometric accuracy in such a constraint, where more related insights are detailed in Sec.~\ref{sec:localize_in_degeneration}. Compared to the reprojection constraint from the LiDAR map as in~\cite{lang2023coco}, the proposed Gaussian-map-based photometric constraint helps better handle situations where LiDAR degeneration and texture
missing occur simultaneously. 
\end{itemize} 

\textbf{Experimentally and practically}, compared to Gaussian-LIC, Gaussian-LIC2 conducts broader experiments with a wider variety of LiDARs of different modalities and sparsity. {\color{blue}Notably}, a specialized LiDAR-Inertial-Camera dataset with ground-truth depths has been curated in Gaussian-LIC2 to enable assessment on out-of-sequence novel view synthesis. Overall, Gaussian-LIC2 achieves state-of-the-art novel-view RGB and depth rendering across various camera trajectories, with Depth L1 reduced by 62\%. Finally, based on Gaussian-LIC2, two applications are presented. The code and the self-collected dataset will be released to benefit the community.

\section{System Overview}
\label{sec:overview}

\begin{figure*}[t!]
    	\centering
    	\includegraphics[width=\textwidth]{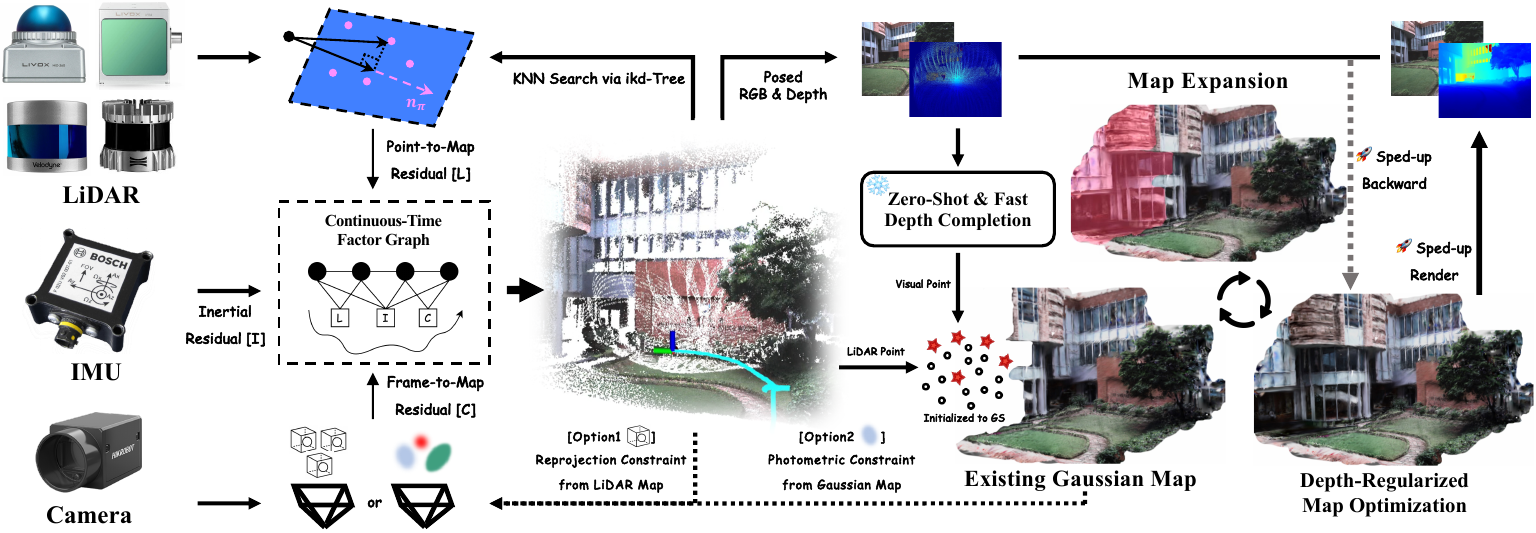}

    	\caption{Pipeline of our real-time photo-realistic LiDAR-Inertial-Camera SLAM which represents the map using 3D Gaussians. 
     }
    \label{fig:pipeline}
\end{figure*}

The notations throughout the paper are given in Tab.~\ref{tab:notation}.
Fig.~\ref{fig:pipeline} depicts the overview of our proposed Gaussian-LIC2, which consists of two main modules: a continuous-time tightly-coupled LiDAR-Inertial-Camera Odometry and an incremental photo-realistic mapping back-end with 3DGS.

In the rest sections, we first present the formulation of continuous-time trajectory in Sec.~\ref{sec:bspline} and introduce the preliminaries of 3DGS in Sec.~\ref{sec:3dgs}. 
Next, in Sec.~\ref{sec:tracking}, we design a tightly-coupled LiDAR-Inertial-Camera odometry system as the front-end which supports two optional camera factors tightly fused within a continuous-time factor graph, including constraints from the Gaussian map. We then utilize an efficient but generalizable depth model to fully initialize Gaussians and prepare mapping data for the back-end in Sec.~\ref{sec:preparation}. Finally, we perform photo-realistic Gaussian mapping with depth regularization and CUDA-related acceleration in Sec.~\ref{sec:3dgs_mapping}.

\begin{table}[t]
\centering
\caption{Notations Glossary.}
\label{tab:notation}
\resizebox{\linewidth}{!}{
\renewcommand{\arraystretch}{1.1} 
\begin{tabular}{cc}
\toprule
Notation & Explanation \\
\midrule
${}^A\mathbf{p}$   & 3D point $\mathbf{p}$ in the frame $\{A\}$ \\
${}^A\mathbf{p}_B$ & translation from frame $\{B\}$ to frame $\{A\}$ \\
${}^A_B\mathbf{R} \in \mathit{SO}(3)$ & rotation matrix from frame $\{B\}$ to frame $\{A\}$ \\
${}^A_B\mathbf{T} \in \mathit{SE}(3)$ & transformation matrix from frame $\{B\}$ to frame $\{A\}$ \\
$\mathbf{p}_i , \mathbf{R}_i$ & positional/rotational control point \\
$\mathbf{\Phi}(t_{\kappa-1}, t_{\kappa})$  & control points of B-splines in $\left[ t_{\kappa-1}, t_{\kappa}\right)$  \\
${}^I\bomega_m, {}^I\mathbf{a}_m$ & angular velocity, linear acceleration in raw IMU measurements \\
$\mathbf{b}_{\omega}^{\kappa-1}, \mathbf{b}_{a}^{\kappa-1}$ & gyroscope and accelerometer biases during  $\left[ t_{\kappa-1}, t_{\kappa}\right)$ \\ 
\midrule
$\boldsymbol{\mu},\boldsymbol{\Sigma}$ & mean and covariance of a 3D Gaussian in frame $\{W\}$ \\
$\mathbf{S} \in \mathbb{R}^3$ & scale of a 3D Gaussian primitive \\
$o \in \mathbb{R}$ & opacity of a 3D Gaussian primitive \\
$\mathbf{SH} \in \mathbb{R}^{3\times 16}$ & spherical harmonics coefficients of a 3D Gaussian primitive\\
$\mathbf{SH}_{b} \in \mathbb{R}^{16}$ & spherical harmonics in a certain viewing direction\\
$\boldsymbol{\mu}^{\prime},\boldsymbol{\Sigma}^{\prime}$ & 2D position and covariance of a 2D Gaussian splatted in a image plane\\
$\mathbf{c}$ & view-dependent color of the 3D Gaussian \\
$\mathbf{\mathcal{G}}$ & Gaussian map in frame $\{W\}$\\
$\mathbf{C}$ & raw RGB image \\
$\hat{\mathbf{C}}(\mathbf{\mathcal{G}}, {}^W_C\mathbf{T}(t))$ & rendered RGB image given camera pose ${}^W_C\mathbf{T}(t)$ \\
\midrule
$\mathbf{D}_s$ & sparse depth map from the latest 5 frames\\
$\mathbf{D}_c$ & completed dense depth map \\
$\mathcal{P}$  & supplemented colorized point cloud selected from unprojected $\mathbf{D}_c$  \\
\bottomrule
\end{tabular}
}
\end{table}

\subsection{Continuous-Time Trajectory Formulation}
\label{sec:bspline}

Two non-uniform cumulative B-splines, parameterizing the 3D rotation and the 3D translation, can jointly represent a continuous-time trajectory. The 6-DoF poses at time $t\in [t_i, t_{i+1})$ of a continuous-time trajectory are denoted by:
\begin{gather}
    \mathbf{R}(t)=F_1\left(\mathbf{R}_{i-3}, \cdots, \mathbf{R}_i, t_i, t_{i+1}, t\right), \label{eq:r-bspline} \\
    \mathbf{p}(t)=F_2\left(\mathbf{p}_{i-3}, \cdots, \mathbf{p}_i, t_i, t_{i+1}, t\right), \label{eq:p-bspline}
\end{gather}
where $\mathbf{R}_n \in \mathit{SO}(3)$ and $\mathbf{p}_n \in \mathbb{R}^{3}$ denote control points ($n \in \{i-3, \cdots,i \}$). $t_i$ and $t_{i+1}$ represent two adjacent knots. The functions $F_1$ and $F_2$ derive poses from control points, knots, and querying time. Please refer to~\cite{lv2023continuous, lang2023coco, lang2022ctrl} for more details. The continuous-time trajectory of IMU in the world frame $\{W\}$ is denoted as ${}^W_I\mathbf{T}(t) = \left[{}^W_I\mathbf{R}(t), {}^W\mathbf{p}_I(t)\right]$. Given known extrinsics between LiDAR/camera and IMU, we can accordingly get LiDAR trajectory ${}^W_L\mathbf{T}(t)$ and camera trajectory ${}^W_C\mathbf{T}(t)$.
\section{3D Gaussian Splatting Representation}
\label{sec:3dgs}

Due to the strengths of 3DGS illustrated in Sec.~\ref{sec:related_work}, we reconstruct a detailed photo-realistic map using a set of anisotropic 3D Gaussians. Each Gaussian is characterized by spatial position $\boldsymbol{\mu} \in \mathbb{R}^3$, scale $\mathbf{S} \in \mathbb{R}^3$, rotation $\mathbf{R} \in \mathbb{R}^{3 \times 3}$, opacity $o \in \mathbb{R}$, and three-degree spherical harmonics coefficients $\mathbf{SH} \in \mathbb{R}^{3 \times 16}$ to encode view-dependent appearance of the scene~\cite{kerbl20233d}. Unlike several prior works~\cite{keetha2024splatam, sun2024mm3dgs}, we do not enforce 3D Gaussians to be isotropic to compromise map quality for speed. Instead, we retain the original expressive parameterization~\cite{kerbl20233d} and adopt the acceleration strategies described in Sec.~\ref{sec:mapping} to pursue both quality and efficiency. 

Representing the Gaussian’s ellipsoidal shape, the covariance of each Gaussian is parameterized as $\boldsymbol{\Sigma} = \mathbf{R}\mathbf{S}\mathbf{S}^T\mathbf{R}^T$. Given a camera pose ${}^C_W\mathbf{T}=\{{}^C_W\mathbf{R}, {}^C\mathbf{p}_W\}$, which maps a 3D point ${}^W\mathbf{p}$ from the world frame $\{W\}$ to the camera frame $\{C\}$, a 3D Gaussian $\mathcal{N}(\boldsymbol{\mu}, \boldsymbol{\Sigma})$ can be splatted onto the image screen, resulting in a corresponding elliptical 2D Gaussian $\mathcal{N}(\boldsymbol{\mu}^\prime, \boldsymbol{\Sigma}^\prime)$:
\begin{gather}
    \boldsymbol{\mu}^{\prime} = \pi_c\left( \frac{\hat{\boldsymbol{\mu}}}{\mathbf{e}_{3}^{\top} \hat{\boldsymbol{\mu}}} \right), \ \hat{\boldsymbol{\mu}} = {}^C_W\mathbf{R} \; \boldsymbol{\mu} + {}^C\mathbf{p}_W\,,
    \\
    \boldsymbol{\Sigma}^{\prime} = \boldsymbol{J} \; {}^C_W\mathbf{R} \; \boldsymbol{\Sigma}  \; {{}^C_W\mathbf{R}}^T\mathbf{J}^T\,,
\end{gather}
where $\mathbf{e}_{i}$ is a $3\times1$ vector with its $i$-th element to be 1 and the other elements to be 0. Thus, ${\mathbf{e}_{3}^{\top} \hat{\boldsymbol{\mu}}}$ gives the depth $d$ of the 3D Gaussian in the camera frame. The function $\pi_c(\cdot)$ projects a 3D point on the normalized image plane to a pixel. $\mathbf{J} \in  \mathbb{R}^{2 \times 3}$ represents the Jacobian of the affine approximation to the perspective projection~\cite{kerbl20233d}.
The projected 2D Gaussian contributes to the image at pixel $\boldsymbol{\rho} = \begin{bmatrix} u & v \end{bmatrix}^\top$ with a weight computed as follows:
\begin{align}
    \alpha = o\exp\left(-\frac{1}{2}(\boldsymbol{\mu}^{\prime}-\boldsymbol{\rho})^T(\boldsymbol{\Sigma}^{\prime})^{-1}(\boldsymbol{\mu}^{\prime}-\boldsymbol{\rho})\right)\,. \label{eq:alpha}
\end{align}

By arranging all the successfully splatted 3D Gaussians in depth order, the color, depth, and opacity at pixel $\boldsymbol{\rho}$ can be efficiently rendered using front-to-back $\alpha$-blending:
\begin{gather}
    \mathbf{C}(\boldsymbol{\rho})=\sum_{i=1}^n\mathbf{c}_i\alpha_i\prod_{j=1}^{i-1}(1-\alpha_j), \label{eq:rendered_rgb} \\
    \mathbf{D}(\boldsymbol{\rho})=\sum_{i=1}^nd_i\alpha_i\prod_{j=1}^{i-1}(1-\alpha_j), 
    \label{eq:rendered_depth}
    \\
\mathbf{O}(\boldsymbol{\rho})=\sum_{i=1}^n\alpha_i\prod_{j=1}^{i-1}(1-\alpha_j), \label{eq:rendered_opacity}
\end{gather}
where $\mathbf{c}$ denotes the view-dependent color of the 3D Gaussian derived from spherical harmonics $\mathbf{SH}_{b} \in \mathbb{R}^{16}$ (as basis functions) and the corresponding coefficients $\mathbf{SH}$ (as weights), while $\mathbf{SH}_{b}$ is computed from the viewing direction between the Gaussian position $\boldsymbol{\mu}$ and camera position ${}^C\mathbf{p}_W$.

To investigate the constraint of the Gaussian map on conventional odometry, we analytically compute the Jacobian of the 3DGS representation w.r.t camera pose on the manifold according to MonoGS~\cite{matsuki2024gaussian}, thereby avoiding the overhead of automatic differentiation. In contrast to MonoGS that models appearance with view-independent color, we further include the Jacobian of high-order spherical harmonics w.r.t the camera pose. The jacobians are as follows:
\begin{gather}
    \label{eq:3dgs_jacobi}
    \frac{\partial \boldsymbol{\mu}^\prime}{\partial{}^C_W\mathbf{T}}=\frac{\partial \boldsymbol{\mu}^\prime}{\partial\hat{\boldsymbol{\mu}}} \frac{\partial \hat{\boldsymbol{\mu}}}{\partial {}^C_W\mathbf{T}}, \nonumber \\
    \frac{\partial \boldsymbol{\Sigma}^\prime}{\partial {}^C_W\mathbf{T}}=\frac{\partial \boldsymbol{\Sigma}^\prime}{\partial \mathbf{J}} \frac{\partial \mathbf{J}}{\partial \hat{\boldsymbol{\mu}}} \frac{\partial \hat{\boldsymbol{\mu}}}{\partial {}^C_W\mathbf{T}}+\frac{\partial \boldsymbol{\Sigma}^\prime}{\partial {}^C_W\mathbf{R}} \frac{\partial {}^C_W\mathbf{R}}{\partial {}^C_W\mathbf{T}}, \\
    \frac{\partial \mathbf{c}}{\partial{}^C_W\mathbf{T}}=\frac{\partial \mathbf{c}}{\partial \mathbf{SH}_{b}}  \frac{\partial\mathbf{SH}_{b}}{\partial{}^C\mathbf{p}_W} \frac{\partial{}^C\mathbf{p}_W}{\partial {}^C_W\mathbf{T}}. \nonumber
\end{gather}
The partial derivative on the manifold is defined as:
\begin{gather}
    \frac{\partial f(\mathbf{T})}{\partial \mathbf{T}} \triangleq \lim _{\boldsymbol{\xi} \rightarrow 0} 
    \frac{ f\big( \mathbf{T \operatorname{Exp}(\boldsymbol{\xi})}\big)  
    - f\big(\mathbf{T}\big)  
    }
    {\boldsymbol{\xi}},
\end{gather}
where $\mathbf{T} \in \mathit{SE}(3)$ and $\boldsymbol{\xi} \in \mathfrak{s e}(3)$. $\operatorname{Exp}(\cdot)$ maps Lie Algebra to Lie Group and $\operatorname{Log}(\cdot)$ is its inverse operation.
\section{LiDAR-Inertial-Camera Fusion}
\label{sec:tracking}

We draw on the insights of classic multi-sensor fused odometry to achieve real-time, robust, and accurate pose estimation, {\color{blue}supporting} for subsequent photo-realistic mapping. Similar to Gaussian-LIC, our odometry system is inherited from and further developed based on our prior work Coco-LIC~\cite{lang2023coco}, a continuous-time tightly-coupled LiDAR-Inertial-Camera odometry using non-uniform B-spline. By leveraging the continuous-time trajectory representation, which inherently supports pose querying at any timestamp corresponding to sensor measurements, we tightly fuse asynchronous, high-frequency LiDAR-Inertial-Camera data without introducing interpolation error. The odometry here supports two types of camera factors, one of which is based on the Gaussian map.

\subsection{Trajectory Extension}
\label{sec:traj_extend}

Our LiDAR-Inertial-Camera system is initialized from a stationary state, using a buffer of IMU measurements to initialize IMU biases and the gravity-aligned orientation~\cite{geneva2020openvins}. After initialization, the system adopts an active sliding window of the latest 0.1 seconds for optimization.

Starting from time instant $t_{\kappa-1}$, we estimate the trajectory in $[t_{\kappa-1}, t_{\kappa})$ once the LiDAR-Inertial-Camera data in the sliding window time interval is ready, where $t_{\kappa} = t_{\kappa-1} + 0.1$. The accumulated data contains all LiDAR raw points $\mathcal{L}_{\kappa}$, all IMU raw data $\mathcal{I}_{\kappa}$, and the latest image frame $\mathcal{F}_{\kappa}$ in $[t_{\kappa-1}, t_{\kappa})$. For simplicity, we omit other image frames captured within this interval, given their negligible impact on the overall performance. As in Coco-LIC, we first adaptively initialize a variable number of control points based on the motion intensity inferred from the IMU data, and then insert these control points into the sliding window to extend the trajectory. The following states are then optimized within the sliding window:
\begin{equation}
\begin{aligned}
    \label{eq:state}
    \mathcal{X}^{\kappa} &= \{\mathbf{\Phi}(t_{\kappa-1}, t_{\kappa}),\, \mathbf{x}_{I_b}^{\kappa}\} , \\
	\mathbf{x}_{I_b}^{\kappa} &= \{ \mathbf{b}_{g}^{\kappa-1},\, \mathbf{b}_{a}^{\kappa-1},\, \mathbf{b}_{g}^{\kappa},\, \mathbf{b}_{a}^{\kappa} \} ,
\end{aligned}
\end{equation}
where $\mathbf{\Phi}(t_{\kappa-1}, t_{\kappa})$ denotes all control points within the interval $[t_{\kappa-1}, t_\kappa)$, parameterizing the continuous-time trajectory of the IMU in the world frame. $\mathbf{x}_{I_b}^{\kappa}$ denotes the IMU bias, which includes the gyroscope bias $\mathbf{b}_g$ and the accelerometer bias $\mathbf{b}_a$. The IMU biases during $[t_{\kappa-1}, t_\kappa)$ are assumed to be constant as $\mathbf{b}_{g}^{\kappa-1}$ and $\mathbf{b}_{a}^{\kappa-1}$. They are under Gaussian random walk and evolve to $\mathbf{b}_{g}^{\kappa}$ and $\mathbf{a}_{g}^{\kappa}$ at $t_\kappa$.

\subsection{Continuous-Time Factor Graph}
\label{sec:factors}
\subsubsection{LiDAR Factor}
Different from Coco-LIC with LiDAR feature extraction and kd-Tree-organized map, we here directly register downsampled LiDAR raw points to the LiDAR map organized by the ikd-Tree for enhanced efficiency~\cite{xu2022fast}. Given a raw LiDAR point ${}^{L}\mathbf{p} \in \mathcal{L}_k$ measured at time $t$, we first transform it to the world frame with the queried pose from the continuous-time trajectory, and then search for its five nearest neighbors in the LiDAR map to fit a 3D plane for a point-to-plane residual:
\begin{gather}
    \label{eq:lidar_factor}
    \mathbf{r}_{L}  = {}^{W}\mathbf{n}^\top_{\pi} {}^{W}\hat{\mathbf{p}} + {}^{W}d_{\pi} \,,\\
    {}^{W}\hat{\mathbf{p}}  = {}^W_L\mathbf{R}(t) {}^L\mathbf{p} + {}^W\mathbf{p}_L(t)\,,
\end{gather}
where ${}^{W}\mathbf{n}_{\pi}$ and ${}^{W}d_{\pi}$ denote the unit normal vector and the distance of the plane to the origin, respectively.

\subsubsection{Inertial Factor}
We define the following inertial factors:
\begin{gather}
    \mathbf{r}_{I}  = \begin{bmatrix} 
    {}^I\bomega(t) - {}^I\bomega_m + \mathbf{b}_{g}^{\kappa-1} \\
    {}^I\mathbf{a}(t) - {}^I\mathbf{a}_m + \mathbf{b}_a^{\kappa-1}
    \end{bmatrix}\,, \\
    \mathbf{r}_{I_b}  = \begin{bmatrix} 
    \mathbf{b}_{g}^{\kappa} - \mathbf{b}_{g}^{\kappa-1} \\
    \mathbf{b}_{a}^{\kappa} - \mathbf{b}_{a}^{\kappa-1}
    \end{bmatrix} \,,
\end{gather}
where the former is the IMU factor and the latter is the bias factor based on the random walk process. ${}^I{\boldsymbol{\omega}_m}, {}^I{\mathbf{a}_m}$ are the raw measurements of angular velocity and linear acceleration of the IMU data at time $t$ in $\mathcal{I}_{\kappa}$. ${}^I\bomega(t)$ and ${}^I\mathbf{a}(t)$ are the corresponding predicted values computed from the derivatives of the continuous-time trajectory in Eq.\eqref{eq:r-bspline} and Eq.\eqref{eq:p-bspline}.

\subsubsection{Camera Factor (Option 1, Default)}
\label{sec:cfactor1}
Our odometry incorporates two types of camera factors, the first of which is based on reprojection constraints using the reconstructed LiDAR map~\cite{lang2023coco}. Specifically, we maintain a subset of global LiDAR points 
stored in voxels and associate them with image pixels by projection, KLT sparse optical flow~\cite{lucas1981iterative}, and RASAC-based
outlier removal. 
Consider a global LiDAR point ${}^W\mathbf{p}$ associated with the pixel $\boldsymbol{\rho} = \begin{bmatrix} u & v \end{bmatrix}^\top$ in the image frame  at the timestamp $t$. The reprojection error for this LiDAR point is defined as:
\begin{align}
    \mathbf{r}_{C_1} = \pi_c\left( \frac{{}^C\hat{\mathbf{p}}}{\mathbf{e}_{3}^{\top} {}^C\hat{\mathbf{p}}} \right) - \begin{bmatrix} u \\ v \end{bmatrix}, \quad 
    {}^C\hat{\mathbf{p}} = {}^W_C\mathbf{T}^{-1}(t) {}^W\mathbf{p}\,.
\end{align}

\subsubsection{Camera Factor (Option 2)}
\label{sec:cfactor2}
The second type of camera factor fully leverages the incrementally constructed Gaussian map. Thanks to the continuous-time trajectory formulation, we can obtain the camera pose ${}^W_{C}\mathbf{T}$ at any time instant within the interval $[t_{\kappa-1}, t_{\kappa})$.
Given the timestamp $t$ at which the raw RGB image $\mathbf{C}$ (treated as ground truth) of frame $\mathcal{F}_{\kappa}$ is captured, we render an RGB image $\hat{\mathbf{C}}(\mathbf{\mathcal{G}}, {}^W_C\mathbf{T}(t))$ from the Gaussian map $\mathbf{\mathcal{G}}$ using Eq.~\eqref{eq:rendered_rgb}. The rendered image is expected to closely match the raw image, enabling optimization of the camera pose by minimizing the loss:
\begin{align}
    \mathcal{L}=\frac{1}{2} \left\|\hat{\mathbf{C}}-\mathbf{C}\right\|_1+\frac{1}{2} \mathcal{L}_{\mathrm{D}-\mathrm{SSIM}}(\hat{\mathbf{C}}, \mathbf{C}),
    \label{eq:tracking_loss}
\end{align}
where $\mathcal{L}_{\mathrm{D}-\mathrm{SSIM}}$ is a D-SSIM term~\cite{kerbl20233d}. Low-image-gradient or low-opacity pixels are penalized~\cite{matsuki2024gaussian}, which is omitted here for simplicity. Besides, in unbounded outdoor scenes with highly variable illumination and under conditions of fluctuating camera exposure, a standalone RGB L1 loss is susceptible to these noises, which may corrupt gradient directions, leading to erroneous camera pose optimization. To this end, we additionally introduce the D-SSIM loss to enhance the optimization robustness by accounting for structural similarity. We perform $N_t$ iterations using the Adam optimizer, during which the Gaussian map $\mathbf{\mathcal{G}}$ is kept fixed while the camera pose is iteratively optimized. The resulting optimized camera pose from Eq.~\eqref{eq:tracking_loss} is denoted as ${}^W_C \widetilde{\mathbf{T}} = \{{}^W_C\widetilde{\mathbf{R}}, {}^W\widetilde{\mathbf{p}}_C\}$, and below is the derived photometric constraint from the Gaussian map:
\begin{align}\label{eq:map_optimize_pose}
    \mathbf{r}_{C_2} = \operatorname{Log}\left({}^W_C\mathbf{T}(t) {}^W_C \widetilde{\mathbf{T}}^{-1}\right).
\end{align}

\subsection{LiDAR-Inertial-Camera Factor Graph Optimization}
\label{sec:lic_factorgraph}
We jointly fuse LiDAR-Inertial-Camera data in the factor graph and formulate the following nonlinear least-squares problem to efficiently optimize states $\mathcal{X}^{\kappa}$:
\begin{equation}
\begin{aligned}
    \label{eq:lvio-graph}
    \arg \min_{\mathcal{X}^{\kappa}}
    \biggl\{ & \sum \left\|\mathbf{r}_L\right\|_{\mathbf{\Sigma}_{L}}^{2}+
    \sum \left\|\mathbf{r}_{I}\right\|_{\mathbf{\Sigma}_{I}}^{2}+
    \sum \left\|\mathbf{r}_{I_b}\right\|_{\mathbf{\Sigma}_{I_b}}^{2}+   \\
    & 
    \sum \left\|\mathbf{r}_{C_x}\right\|_{\mathbf{\Sigma}_{C}}^{2}+
    \sum \left\|\mathbf{r}_{\text{prior}}\right\|_{\mathbf{\Sigma}_{\text{prior}}}^{2} \biggr\}
    \,,
\end{aligned}
\end{equation}
which is solved via the Levenberg-Marquardt algorithm in Ceres Solver~\cite{ceres-solver} and accelerated through the analytical derivatives. The camera factor $\mathbf{r}_{C_x}$ can be chosen as either $\mathbf{r}_{C_1}$ or $\mathbf{r}_{C_2}$. The former is a reprojection constraint from the LiDAR map, which is somewhat handcrafted but very lightweight, while the latter is a photometric constraint from the Gaussian map, which consumes modest GPU computing resources but is more direct and natural. A comparative evaluation of the two factors will be presented in Sec.~\ref{sec:experiment}. $\mathbf{r}_{\text{prior}}$ is the prior factor from marginalization~\cite{lang2023coco}. $\mathbf{\Sigma}_{L}$, $\mathbf{\Sigma}_{I}$, $\mathbf{\Sigma}_{I_b}$, 
$\mathbf{\Sigma}_{C}$,
$\mathbf{\Sigma}_{\text{prior}}$ are the corresponding measurement covariances.
\section{Data Preparation for Gaussian Mapping}
\label{sec:preparation}

\subsection{Mapping Data Grouping}
\label{sec:grouping}
After finishing the optimization of the latest trajectory segment in $[t_{\kappa-1}, t_{\kappa})$, we ultimately obtain the accurate camera pose for the latest image frame $\mathcal{F}_{\kappa}$. All LiDAR raw points within this time interval, namely $\mathcal{L}_{\kappa}$, can be easily transformed into the world frame based on the optimized continuous-time trajectory. The posed image, together with the transformed LiDAR points, is jointly regarded as a hybrid frame, and we treat every fifth hybrid frame as a keyframe for photo-realistic mapping. If the current hybrid frame is selected as a keyframe, we merge all LiDAR points from the latest five hybrid frames into a single point cloud. This point cloud is then projected onto the current image plane to generate a sparse depth map. Next, we downsample the point cloud by randomly retaining one out of every $N_p$ points. The downsampled point cloud is also projected onto the current image plane for coloring. 
Colorzied points will be used for initializing 3D Gaussians, which will be discussed in Sec.~\ref{sec:expansion}. At this stage, the keyframe is fully constructed, consisting of a posed image, a sparse depth map, and a set of colorized LiDAR points.

\subsection{LiDAR Blind Area Compensation}
\label{sec:dc}
\begin{figure}[t!]
    \centering
    \includegraphics[width=\linewidth]{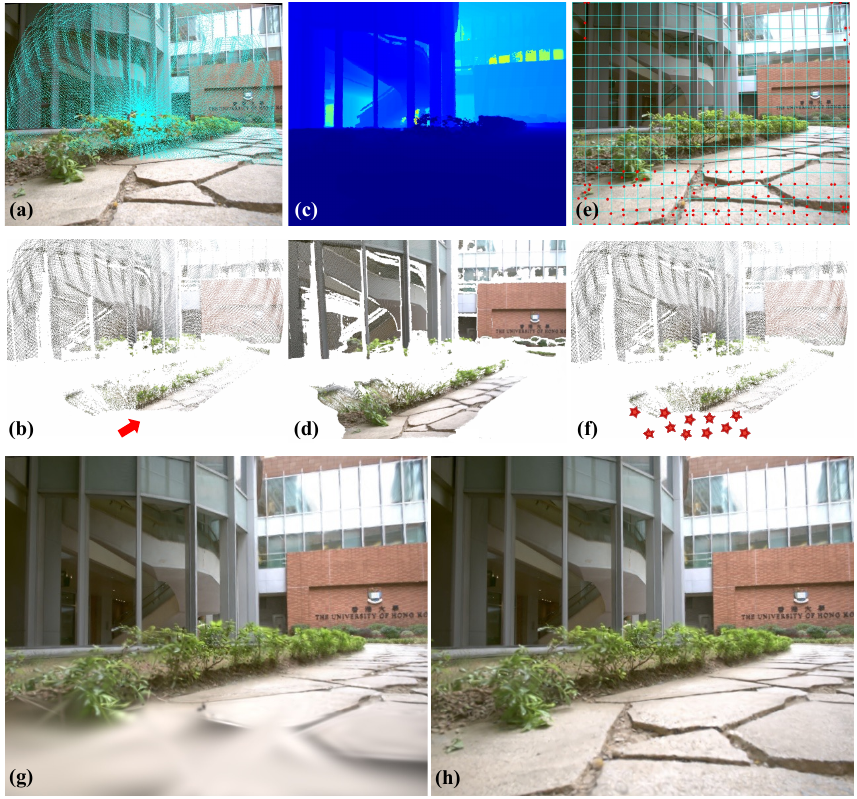}
    \caption{Details on the LiDAR blind area compensation: (a) Projection of LiDAR points of the latest 5 frames. (b) Colorized LiDAR points of the latest 5 frames. It is worth noting that the area highlighted by the red arrow has been never scanned by the LiDAR during the entire data acquisition. (c) Completed dense depth map. (d) Dense colored point cloud from the completed depth map. (e) The image is divided into $30 \times 30$ patches, and the red dot denotes the selected pixels. (f) The red star indicates the 3D visual points obtained by back-projecting the selected pixels to compensate for the LiDAR blind area. (g) and (h) show the qualitative rendering results of Gaussian-LIC2 without and with LiDAR blind area compensation.}
    \label{fig:dc}
\end{figure}

LiDARs are able to provide precise geometric priors for Gaussian initialization. However, due to the mismatch in FoV between the LiDAR and the camera, relying solely on LiDAR points would lead to under-reconstruction in LiDAR blind areas or when using the sparse LiDAR. Although the adaptive density control (ADC) from vanilla 3DGS~\cite{kerbl20233d} offers a partial relief, its effectiveness is limited in SLAM. The incremental insertion of Gaussians leads to inconsistent convergence among different batches of Gaussians, which may misguide gradient-based operations like cloning and splitting. Moreover, the reliance on multi-iteration gradient accumulation hinders the real-time performance of the system. Gaussian-LIC leverages online triangulated visual features to compensate for regions unobserved by LiDAR. However, it struggles in textureless areas and is sensitive to inter-frame viewpoint variations. Therefore, further investigation is warranted to explore how visual cues can be more effectively leveraged for efficient Gaussian initialization. The central challenge lies in obtaining accurate depth for image pixels.

\textbf{Why should it be depth completion?} With the advancement of computer vision research, numerous methods have emerged for recovering dense depth from images. First, monocular relative depth estimation~\cite{yang2024depth_v2, wang2024moge} has made significant progress in predicting depth from a single RGB image. However, it inherently suffers from scale ambiguity. While least-squares alignment with metric LiDAR depth can be applied, it often fails to recover accurate metric scale across all image pixels—particularly in regions with poor relative depth estimation or lacking LiDAR coverage~\cite{zuo2025dvgt}. Second, monocular metric depth estimation~\cite{hu2024metric3d, bochkovskii2024depth} directly predicts scale-aware depth. However, its accuracy remains modest and lags far behind LiDAR-based methods, especially in large-scale outdoor environments. Third, multi-view stereo (MVS) approaches~\cite{cao2022mvsformer, xu2023unifying} estimate metric depth from multiple posed images. Although MVS can produce accurate depth in favorable conditions, it is sensitive to parallax and exhibits limited generalization, particularly in large-scale outdoor scenes. In contrast, depth completion, which fuses sparse LiDAR data with a single RGB image, offers a practical and robust solution for recovering metric dense depth~\cite{liu2024depthlab, lin2024prompting} in large-scale scenarios. It avoids the issue of scale ambiguity and can achieve LiDAR-comparable depth estimation accuracy.

In this paper, SPNet~\cite{wang2024scale} is selected as the depth completer due to its efficiency, compactness, and strong generalization ability. We use it directly off the shelf, without any additional fine-tuning. As outlined in Alg.~\ref{alg:completion} and illustrated in Fig.~\ref{fig:dc}, we feed the sparse depth map $\mathbf{D}_s$ and the RGB image $\mathbf{C}$ of the keyframe in Sec.~\ref{sec:grouping} into SPNet to generate a completed dense depth map $\mathbf{D}_c$. We then compute the mean depth change in the known regions before and after completion. If the depth change exceeds a threshold $\epsilon_1$, the completion is considered a failure (rarely occurring). After successful completion, pixels in $\mathbf{D}_c$ with negative depth values or high depth gradient magnitudes are discarded, producing the filtered dense depth map $\mathbf{D}_{c_f}$. Subsequently, we divide the input sparse LiDAR depth map into patches of size $30 \times 30$ and iterate over each of them. For patches without any valid LiDAR depth, we select the pixel with the smallest completed depth within the patch of $\mathbf{D}_{c_f}$ and store it in a container $\mathcal{A}$ if its depth is less than $\epsilon_2$. All pixels in $\mathcal{A}$ are back-projected to form a supplemented colorized point cloud $\mathcal{P}$, compensating for LiDAR-unobserved regions. Finally, $\mathcal{P}$ is transformed to the world frame and merged with the colored LiDAR points of the keyframe.

\begin{algorithm}[t]
    \DontPrintSemicolon
    \caption{LiDAR Blind Area Compensation}
    \label{alg:completion}
    
    \KwIn{Sparse depth map $\mathbf{D}_s$, RGB image $\mathbf{C}$}
    \KwOut{Supplemented colored point cloud $\mathcal{P}$}
    
    $\mathcal{P} \leftarrow \emptyset$, $\mathcal{A} \leftarrow \emptyset$ 
    \tcp*{Initialize outputs}

    $\mathbf{M} \leftarrow \texttt{ValidMask}(\mathbf{D}_s)$ \;

    $\mathbf{D}_c \leftarrow \texttt{SPNet}(\mathbf{C}, \mathbf{D}_s, \mathbf{M})$ \;

    $\delta \leftarrow \texttt{Mean}(|(\mathbf{D}_c - \mathbf{D}_s)[\mathbf{M}]|)$ \;

    \If{$\delta < \epsilon_1$}{
        $\mathbf{D}_{c_f} \leftarrow \texttt{Filter}(\mathbf{D}_c)$ \;

        Divide $\mathbf{D}_s$ into $30 \times 30$ grid patches \;

        \ForEach{patch $\mathcal{R}$ in grid}{
            \If{$\texttt{IsEmpty}(\mathbf{D}_s[\mathcal{R}])$}{
                $\rho \leftarrow \texttt{MinValidDepthPixel}(\mathbf{D}_{c_f}[\mathcal{R}])$\;
                \If{$\texttt{Depth}(\rho) < \epsilon_2$}{
                    $\mathcal{A} \leftarrow \mathcal{A} \cup \{\rho\}$\;
                }
            }
        }
        $\mathcal{P} \leftarrow \texttt{BackProject}(\mathcal{A}, \mathcal{I})$\;
    }
    \Return{$\mathcal{P}$}
\end{algorithm}

\section{Real-Time Photo-Realistic Mapping}
\label{sec:3dgs_mapping}

The mapping back-end thread runs in parallel with the tracking front-end thread and continuously receives sequential keyframes from the tracker to perform real-time photo-realistic reconstruction. Each keyframe contains an estimated camera pose, an undistorted image, a sparse LiDAR depth map, and a set of colored LiDAR points augmented with supplemented visual points to compensate for the blind area.

\subsection{Gaussian Map Management}
\label{sec:gaussian_map_management}
Once a keyframe is received, the mapping thread will initialize or expand the Gaussian map and optimize it.

\subsubsection{Initialization of Gaussian Map}
\label{sec:map_init}
The Gaussian map is initialized using all the colorized LiDAR points and the supplemented visual points from the first keyframe. 
Specifically, for each point, a new Gaussian is instantiated at its 3D location, with the zeroth degree of $\mathbf{SH}$ initialized using its RGB color, opacity set to 0.1, and rotation initialized as the identity matrix.
To mitigate aliasing artifacts~\cite{yan2024multi}, we adapt the scale of each Gaussian
based on its distance to the image plane. The scale is modeled as $\mathbf{S} = \frac{d}{f} \mathbf{e}$, where $\mathbf{e}$ is a $3 \times 1$ vector of ones, $d$ is the depth of the 3D point in the camera frame, and $f$ is the focal length.

\subsubsection{Expansion of Gaussian Map}
\label{sec:expansion}
Each incoming keyframe typically captures new geometric and appearance information. 
To maintain efficiency, we focus on initializing new Gaussians for the newly observed parts of the scene.
Specifically, we first render an opacity map $\mathbf{O}$ from the perspective of the current keyframe using Eq.~\eqref{eq:rendered_opacity}. A binary mask $\mathbf{M}_{o} = \mathbf{O} < \tau$  ($\tau$ is a constant threshold) is then constructed to identify potentially newly observed image regions. 
Finally, only the 3D points projected into these image regions are selected to initialize new Gaussians.

\subsubsection{Depth-Regularized Map Optimization}
\label{sec:map_opt}
After the initialization or expansion, we randomly sample $K$ keyframes out of all keyframes to optimize the Gaussian map, preventing catastrophic forgetting and ensuring global geometric consistency. The selected keyframes are shuffled and sequentially used to update the map by minimizing the following rendering loss:
\begin{gather}
    \mathcal{L}=\mathcal{L}_c+\gamma\mathcal{L}_d, \\
    \mathcal{L}_{c}=(1-\lambda) \left\|\hat{\mathbf{C}}-\mathbf{C}\right\|_1+\lambda \mathcal{L}_{\mathrm{D}-\mathrm{SSIM}}(\hat{\mathbf{C}}, \mathbf{C}), \\
    \mathcal{L}_d=\left\|\left(\frac{\hat{\mathbf{D}}}{\hat{\mathbf{O}}}-{\mathbf{D}_s}\right) \cdot \mathbf{M}_d\right\|_1, \ \mathbf{M}_d={\mathbf{D}_s}>0, \label{eq:depth_loss}
\end{gather}
where $\mathcal{L}_c$ denotes the RGB rendering loss computed by the rendered image $\hat{\mathbf{C}}$ based on Eq.\eqref{eq:rendered_rgb} and the raw 
image $\mathbf{C}$, while $\mathcal{L}_d$ is the depth rendering loss involved with the rendered depth map $\hat{\mathbf{D}}$ via Eq.\eqref{eq:rendered_depth} and the sparse LiDAR depth map ${\mathbf{D}_s}$. Note that the incremental insertion of Gaussians in SLAM can cause the rendered depth of certain pixels to be significantly smaller than the ground-truth values. We here normalize the rendered depth map using the rendered opacity map $\hat{\mathbf{O}}$ to correct the depth supervision.

\subsection{CUDA-Related Acceleration Strategies}
\label{sec:acc}
To improve real-time performance, previous works either forcibly limit the number of Gaussians and optimization iterations or simplify the Gaussian representation~\cite{keetha2024splatam, peng2024rtg}, which might work well indoor but often degrades performance in unbounded and complex outdoor scenarios. On the contrary, we concentrate on speeding up the CUDA-based operations both in terms of algorithms and engineering, particularly the forward and backward propagation of the 3DGS rasterizer, so as to ensure real-time performance without compromising the quality.

\subsubsection{Fast Tile-based Culling}
\begin{figure}[t!]
    \centering
    \includegraphics[width=\linewidth]{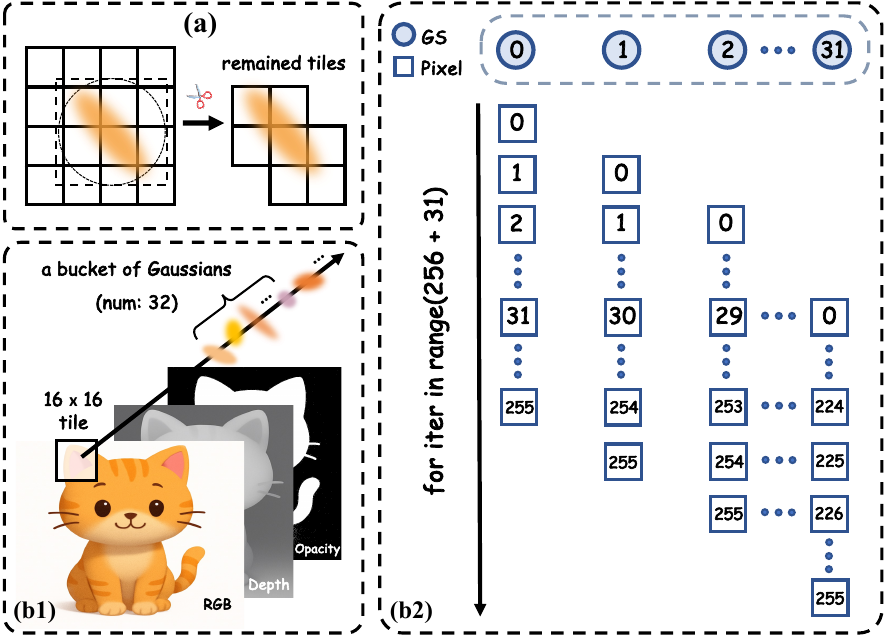}
    \caption{Acceleration strategies in the forward and backward pass when optimizing the Gaussian map: (a) Culling the tiles weakly affected by the splatted Gaussians. (b1) Every 32 Gaussians within a tile ($16 \times 16$ pixels) are grouped into a bucket. (b2) All buckets are processed in parallel for (256+31) iterations (row by row). In each iteration, all Gaussians within the bucket accumulate gradients from different pixels for optimization with respect to the rendered RGB, depth, and opacity maps—without incurring atomic collisions.}
    \label{fig:cuda}
\end{figure}

During the forward pass, 3D Gaussians are splatted onto the image, resulting in elliptical 2D Gaussians. To determine the tiles affected by each Gaussian, the elliptical 2D Gaussian is further dilated into a circle whose radius equals the major axis length~\cite{kerbl20233d}. However, such an approximation results in overly inflated tiles influenced by the Gaussian, particularly for highly anisotropic Gaussians, which are common in incremental mapping systems such as SLAM. To address this issue, we adopt a fast tile-based culling strategy, as illustrated in Fig.~\ref{fig:cuda}(a). For each 2D Gaussian $\mathcal{N}(\boldsymbol{\mu}^\prime, \boldsymbol{\Sigma}^\prime)$ and the tiles affected by it, we identify the pixel $\rho$ within every tile where the 2D Gaussian yields the highest contribution (with the max weight $\alpha$ in Eq.\eqref{eq:alpha}), as in ~\cite{radl2024stopthepop}. If the weight $\alpha$ at the pixel $\rho$ is smaller than the threshold $\frac{1}{255}$, we regard the tile as weakly affected and cull it. In this way, the number of Gaussians per tile can be significantly reduced, accelerating the subsequent forward and backward pass.

\subsubsection{Per-Gaussian Backpropagation}
During the backward pass, gradients flow from RGB pixels back to Gaussians, which is the most time-consuming stage during the mapping~\cite{kerbl20233d}. Each pixel typically receives contributions from a different number of Gaussians, and the runtime is ultimately dominated by the pixel associated with the largest number of Gaussians. As a result, the remaining CUDA threads must wait idly, leading to significant inefficiency. Also, atomic gradient addition
conflicts arise when multiple pixels backpropagate to the same Gaussian. To address these issues, we shift from pixel-wise parallelism to Gaussian-wise parallelism following ~\cite{mallick2024taming}, as illustrated in Fig.~\ref{fig:cuda}(b1) and (b2). To be specific, we divide the depth-sorted Gaussians within each tile into buckets of 32 Gaussians each. Then, all Gaussians across all buckets are processed in parallel. For a given bucket, all 32 Gaussians simultaneously iterate over pixels within the tile to accumulate gradients. As a result, the upper bound of the backward time is largely determined by the number of pixels, leading to more consistent computational load and reduced collisions. 
In addition to the RGB channels, we also incorporate per-Gaussian gradient backpropagation for the depth map and the opacity map, which has not been handled in Gaussian-LIC.

\subsubsection{Additional Strategies}
\label{sec:add_strategy}
\textit{\textbf{a) Sparse Adam.}} \ The original 3DGS~\cite{kerbl20233d} uses the Adam optimizer to update all Gaussians, including those not involved in the current rendering, by applying zero gradients. The computational burden becomes progressively unsustainable as the map size increases. Therefore, we adopt sparse Adam~\cite{mallick2024taming} to update only the valid Gaussians that participate in the rendering. \, \textit{\textbf{b) Separated $\mathbf{SH}$.}} \ In the original 3DGS framework, the low-order and high-order $\mathbf{SH}$ coefficients are concatenated before the forward and backward passes per iteration. However, the concatenation operation is time-consuming, especially when using three-degree $\mathbf{SH}$ in our method. To address this, we handle the low-order and high-order coefficients separately, eliminating the concatenation. \, \textit{\textbf{c) Efficient SSIM.}} \ We employ the highly efficient and differentiable SSIM implemented in~\cite{mallick2024taming}, where separable Gaussian convolutions lead to reduced computation. \, \textit{\textbf{d) Warp-Level Predicate Masking.}} \ The warp serves as a bridge between threads and blocks, which is crucial for parallel performance in GPU. Threads are typically operated in groups of 32 as a warp, and performance can degrade if threads within a warp diverge due to conditional branches. For example, in the forward pass of 3DGS,
Gaussians handled by different threads may be deemed invalid—due to invalid depth, non-invertible 2D covariance, or insufficient weights—causing threads to exit under varying conditions, leading to divergence. To tackle this problem, we use flags to unify the return decisions in the final stage, avoiding divergence. Besides, we implement early stopping for warps in which all threads are flagged to return, improving efficiency. \, \textit{\textbf{e) CPU to GPU.}} \ When optimizing the map, ground-truth images and sparse LiDAR depth maps need to be transferred from the CPU to the GPU, which incurs a certain amount of overhead. However, preloading them onto the GPU is not allowed due to limited memory. Thus, we store them in pinned memory and utilize non-blocking transfer to move data from the CPU to the GPU. Specifically, the pinned memory is a special memory region on the CPU that is directly accessible by the GPU, which accelerates the data transfer. The \texttt{cudaMemcpyAsync()} function is used to perform the transfer asynchronously, while CUDA streams are employed to manage concurrent execution of memory copy operations and GPU kernels. By utilizing multiple streams, data transfer and computation can occur simultaneously, minimizing idle time for the GPU and improving overall throughput.

\section{Datasets for Evaluation}
\label{sec:data}

This section describes the datasets used for evaluation, encompassing both public and self-collected ones. The public datasets, including R3LIVE~\cite{lin2022r}, FAST-LIVO~\cite{zheng2022fast, zheng2024fast}, MCD~\cite{nguyen2024mcd}, and M2DGR~\cite{yin2021m2dgr}, are mainly employed for localization and in-sequence novel view rendering evaluation. Our self-collected dataset supports both in-sequence and out-of-sequence novel view rendering evaluation, and enables quantitative evaluation of rendered depth. Tab.~\ref{tab:runtime} summarizes the duration and length of each sequence.

\subsection{Public Dataset}

The R3LIVE dataset~\cite{lin2022r} and FAST-LIVO dataset~\cite{zheng2022fast, zheng2024fast} are both collected within the campuses using a handheld device
equipped with a Livox Avia LiDAR at 10 Hz and its built-in IMU at 200 Hz, and a 15 Hz RGB camera. Closely aligned with the camera’s FoV and performing non-repetitive scanning, the projected points of the adopted solid-state LiDAR are distributed across the image, while still suffering from non-negligible blind spots as shown in Fig.~\ref{fig:dc}. 
Six outdoor sequences from the R3LIVE dataset that are well-suited for mapping are selected for evaluation. The FAST-LIVO dataset used in this work includes all sequences from FAST-LIVO and the first two sequences from FAST-LIVO2, whose high-quality images make the dataset particularly suitable for mapping tasks.
Notably, the R3LIVE and FAST-LIVO datasets also provide challenging sequences with LiDAR or visual degeneration, such as \textit{degenerate\_seq\_00}, \textit{degenerate\_seq\_01}, \textit{LiDAR\_Degenerate}, and \textit{Visual\_Challenge}. Although the ground-truth trajectory is unavailable, the start and end poses coincide, allowing localization evaluation via start-to-end drift.
When stereo pairs are available, only the left image ($640\times512$) is used.

The MCD dataset~\cite{nguyen2024mcd} provides sequences captured in large-scale campuses. We here pick the sequences (\textit{tuhh\_day\_02-04}) with the repetitive mechanical spinning Ouster OS1-64 LiDAR at 10Hz,  a 30 Hz RGB camera ($640\times480$) and a 400 Hz IMU. 
The spatial sparsity of spinning LiDAR poses greater challenges for LiDAR-based Gaussian mapping.
Due to the extensive scale of the environment, we truncate the sequences to the first 200 seconds ($\sim$300 m). The dataset provides highly accurate ground-truth trajectories for quantitative evaluation of localization.

The M2DGR dataset~\cite{yin2021m2dgr}, collected by a ground robot using a Velodyne VLP-32C LiDAR at 10 Hz, an IMU at 100 Hz, and a 10 Hz RGB camera ($640\times480$), presents a more challenging case due to the extremely sparse LiDAR measurements. A set of sequences (\textit{room\_01-03}) under favorable lighting are selected for evaluation.

\subsection{Self-Collected Dataset}

\begin{figure}[t!]
    \centering
    \includegraphics[width=\linewidth]{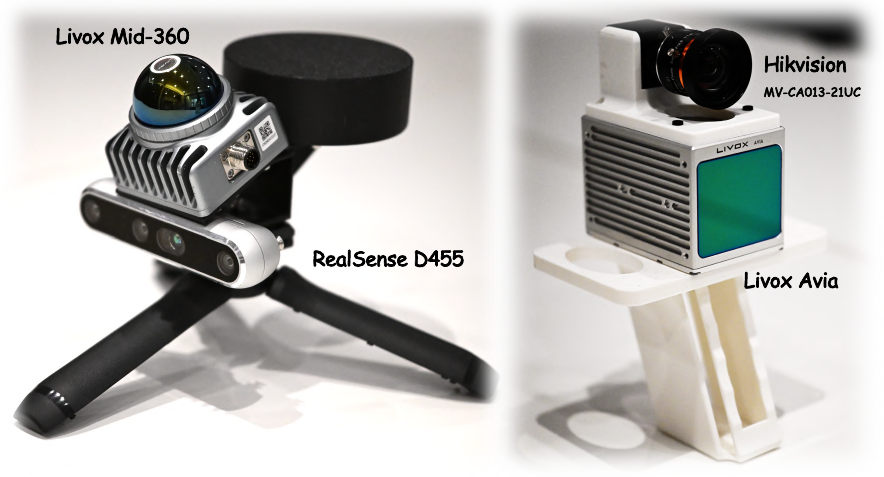}
    \caption{Two types of LIC sensor rig for self-collected dataset.}
    \label{fig:setup}
\end{figure}

To further evaluate out-of-sequence novel view synthesis and quantitatively assess the accuracy of the rendered depth, 
we collect additional sequences leveraging two handheld devices in Fig.~\ref{fig:setup}, enriching LiDAR modality diversity. The first setup integrates a Livox Mid-360 LiDAR
sampled at 10 Hz paired with a 30 Hz RealSense D455 RGB camera ($640\times480$). Compared to Avia, Mid-360 offers a wider horizontal FoV but shorter maximum ranging distance, presenting unique reconstruction challenges. Meanwhile, the second one follows the LIV\_handheld \footnote{\url{https://github.com/xuankuzcr/LIV_handhold}} configuration, combining a Livox Avia LiDAR at 10 Hz with a Hikvision MV-CA013-21UC RGB camera at 15 Hz ($640\times512$). Both handheld systems utilize the built-in 200 Hz IMU from the LiDAR, with the LiDAR and the IMU factory-synchronized and well-calibrated. The spatial and temporal extrinsics between the camera and the IMU are carefully calibrated through~\cite{furgale2013unified}.

For each sequence, we capture a long and smooth trajectory with loop closures, ensuring the scene is revisited for out-of-sequence evaluation and making the captured data well-suited for mapping purposes. 
Due to unreliable GPS between buildings,
GPS-based pose estimation is often unreliable. We generate continuous-time ground-truth trajectories using a LiDAR-Inertial-Camera SLAM framework that incorporates continuous-time loop closure~\cite{lv2021clins} to enhance accuracy. 
Note that our captured scenes are not excessively large (i.e., not at an urban scale), and the estimated ground-truth trajectories are sufficiently accurate for evaluation purposes. For generating ground-truth depth maps, we query the camera pose from the ground-truth continuous-time trajectory and project the temporally closest LiDAR frame onto the image. Each collected trajectory is divided into two segments, used respectively for in-sequence and out-of-sequence evaluations, as illustrated in Fig.~\ref{fig:out-of-seq}.

\section{Experiment Results}
\label{sec:experiment}

\subsection{Implementation Details}
\label{sec:impl_detail}
We implement Gaussian-LIC2 fully in C++ and CUDA. The tracking and mapping modules run in parallel and communicate via ROS, with the mapping built upon the LibTorch framework. The depth completion model~\cite{wang2024scale} is deployed with TensorRT for efficient inference.

For odometry, we reuse the ikd-Tree parameters from FAST-LIO2~\cite{xu2022fast}, set the camera pose optimization step $N_t$ in Sec.~\ref{sec:cfactor2} to 30. 
For mapping data preparation, we set $N_p$ in Sec.~\ref{sec:grouping} to 10 (Avia, Mid-360), 5 (OS1-64), and 1 (VLP-32C). $\epsilon_1$ and $\epsilon_2$ in Sec.~\ref{sec:dc} are empirically set to 0.1 m and 50 m, respectively. For map expansion, we set $\tau$ in Sec.~\ref{sec:expansion} to 0.99 as in Gaussian-LIC. As for map optimization, we set the loss weighting $\lambda$ to 0.2 as in~\cite{kerbl20233d} and $\gamma$ to 0.005. $\gamma$ is determined through rigorous adjustment to strike a balance between visual and geometric quality. The number $K$ of selected keyframes is set to 100. All the learning rates for Gaussian attributes follow the vanilla 3DGS~\cite{kerbl20233d} but do not decay with schedulers. We keep the same hyperparameters in all sequences with the same sensor setup to ensure a fair and comprehensive evaluation. Experiments are run on a desktop PC with an NVIDIA RTX 4090 GPU (24 GB VRAM), Intel i9-13900KF CPU, and 64 GB RAM.

\subsection{Baselines}
This paper focuses on developing an elaborated radiance-filed-based SLAM system capable of robust and accurate pose estimation while constructing photo-realistic 3D maps in real time. Therefore, our comparative analysis specifically targets authentic SLAM systems, explicitly excluding pure mapping systems that require a priori ground-truth poses.

We first evaluate localization performance against state-of-the-art traditional LiDAR-Inertial-Camera SLAM systems, including optimization-based LVI-SAM~\cite{shan2021lvi} and filter-based R3LIVE~\cite{lin2022r} and FAST-LIVO2~\cite{zheng2024fast}. We also include two neural SLAM methods: PIN-SLAM~\cite{pan2024pin}, a neural LiDAR SLAM approach utilizing point-based NeRF for geometry-only reconstruction, and DBA-Fusion~\cite{zhou2024dba}, which integrates IMU data into DROID-SLAM~\cite{teed2021droid}, currently the most advanced neural visual SLAM framework. We run PIN-SLAM by feeding in undistorted LiDAR point clouds.

Furthermore, to evaluate both the localization and photo-realistic mapping performances of the entire system, we compare against photo-realistic radiance-field-based SLAM, including both NeRF-based and 3DGS-based. 
1) \textbf{NeRF-based}: As no related LiDAR-Camera system is publicly available, we adapt the state-of-the-art RGB-D NeRF-based method Co-SLAM~\cite{wang2023co} with pseudo RGB-D input by merging LiDAR scans.
When rendering the full image, we use SPNet~\cite{wang2024scale} to complete the LiDAR depth maps for depth-guided sampling. 
2) \textbf{3DGS-based}: 
We compare with the state-of-the-art RGB-only method MonoGS~\cite{matsuki2024gaussian}, which elegantly derives pose gradients. 
Given the scarcity of available LiDAR-Camera 3DGS-based methods, we run the state-of-the-art RGB-D 3DGS-based approach SplaTAM~\cite{keetha2024splatam} using pseudo RGB-D images.
We also compare against the RGB-D inertial method, MM3DGS-SLAM~\cite{sun2024mm3dgs}, which extends SplaTAM by incorporating IMU data, and include a comparison with our previous LiDAR-Inertial-Camera 3DGS-based SLAM framework Gaussian-LIC~\cite{lang2025gaussian}, where the sky modeling is disabled to enable valid depth rendering. Additional comparison results, including those against the recent LiDAR-Inertial-Camera-based method GS-LIVM~\cite{xie2024gs}, are provided in the supplementary material (Sec.~\ref{sec:gslivm}). For fairness, all methods are evaluated without loop closure and post-processing.

\subsection{Experiment-1: Evaluation of Localization}
\label{sec:localization}
We validate the robustness and accuracy of Gaussian-LIC2 tracking in both challenging degenerate and large-scale environments. In particular, we compare two types of camera factors presented in Sec.~\ref{sec:factors}: option 1 (LiDAR map reprojection constraint, Gaussian-LIC2 (c1)) and option 2 (Gaussian map photometric constraint, Gaussian-LIC2 (c2)). Also, we investigate the effectiveness of the camera factor option 2 when the depth supervision (Eq.~\eqref{eq:depth_loss}) is disabled, denoted as Gaussian-LIC2 (c2-w/o-d). Gaussian-LIC2-LIO represents our LiDAR-Inertial odometry component without any constraints from images.

\begin{table*}[t]
\centering
\caption{Localization Evaluation: The start-to-end drift error (translation $|$ rotation) on challenging sequences. The best are in bold and the second best are \underline{underscored}. $\times$ means failure. Note that, except for the sequence \textit{LiDAR\_Degenerate}, which is an indoor sequence, all the others are outdoor sequences. Hybrid means using both the point cloud map and the 3DGS map for localization.}
\label{tab:challenging_seq_comparison}
\resizebox{\linewidth}{!}{
\begin{tabular}{
    lcc
    c@{\hspace{6pt}{$\;|\;$}\hspace{6pt}}c  
    c@{\hspace{6pt}{$\;|\;$}\hspace{6pt}}c  
    c@{\hspace{6pt}{$\;|\;$}\hspace{6pt}}c  
    c@{\hspace{6pt}{$\;|\;$}\hspace{6pt}}c  
}
\toprule
\multirow{2}{*}{Method} &
\multirow{2}{*}{Sensor} &
\multirow{2}{*}{\begin{tabular}[c]{@{}c@{}}Map Type\\ (for Loc)\end{tabular}} &
\multicolumn{8}{c}{Sequence} \\
\cmidrule(lr){4-11}
& & &
\multicolumn{2}{c}{degenerate\_seq\_00} &
\multicolumn{2}{c}{degenerate\_seq\_01} &
\multicolumn{2}{c}{LiDAR\_Degenerate} &
\multicolumn{2}{c}{Visual\_Challenge} \\
\midrule
MonoGS~\cite{matsuki2024gaussian}       & C         & 3DGS       & 14.54m  & 67.40° & \multicolumn{2}{c}{$\times$} & \multicolumn{2}{c}{$\times$} & \multicolumn{2}{c}{$\times$} \\
Co-SLAM~\cite{wang2023co}      & L + C     & NeRF       & 12.37m & 30.50° & \multicolumn{2}{c}{$\times$} & \multicolumn{2}{c}{$\times$} & \multicolumn{2}{c}{$\times$} \\
SplaTAM~\cite{keetha2024splatam}      & L + C     & 3DGS       & 17.96m & 16.75° & \multicolumn{2}{c}{$\times$} & \multicolumn{2}{c}{$\times$} & \multicolumn{2}{c}{$\times$} \\
MM3DGS-SLAM~\cite{sun2024mm3dgs}  & L + I + C & 3DGS       & 9.14m & 14.02° & \multicolumn{2}{c}{$\times$} & \multicolumn{2}{c}{$\times$} & \multicolumn{2}{c}{$\times$} \\
PIN-SLAM~\cite{pan2024pin}     & L         & NeRF       & 9.07m & 30.05° & 13.53m & 29.29° & 4.13m & 3.71°  & \multicolumn{2}{c}{$\times$} \\
DBA-Fusion~\cite{zhou2024dba}   & I + C     & PointCloud & 0.53m & 8.98°  & 1.87m  & 1.93°  & 0.46m & 4.66°  & 3.19m & 6.21° \\
LVI-SAM~\cite{shan2021lvi}      & L + I + C & PointCloud & 0.08m & 6.60°  & 0.11m  & 2.67°  & \multicolumn{2}{c}{$\times$} & \multicolumn{2}{c}{$\times$} \\
R3LIVE~\cite{lin2022r}       & L + I + C & PointCloud & \underline{0.04m} & \textbf{0.41°} & 0.11m  & \textbf{0.55°} & 8.52m & 2.99°  & 0.21m & 0.69° \\
FAST-LIVO2~\cite{zheng2024fast}   & L + I + C & PointCloud & 5.07m  & 8.38°  & 2.41m  & 6.20°  & \textbf{0.02m} & \textbf{2.40°} & \textbf{0.02m} & \underline{0.14°} \\
Gaussian-LIC~\cite{lang2025gaussian}       & L + I + C & PointCloud & \underline{0.04m} & 0.54° & 0.06m & 0.62° & \underline{0.05m} & 2.63° & 0.07m & 0.30° \\
Gaussian-LIC2-LIO & L + I & PointCloud & 7.87m & 9.15° & 13.72m & 11.10° & 6.62m & 4.84° & 12.05m & 1.41° \\
Gaussian-LIC2 (c1)       & L + I + C & PointCloud & \underline{0.04m} & 0.58° & \textbf{0.04m} & \textbf{0.55°} & \underline{0.05m} & 2.58° & \underline{0.06m} & 0.23° \\
Gaussian-LIC2 (c2)       & L + I + C & Hybrid     & \textbf{0.03m} & \underline{0.43°} & \underline{0.05m} & \underline{0.59°} & \underline{0.05m} & \underline{2.50°} & \textbf{0.02m} & \textbf{0.11°} \\
Gaussian-LIC2 (c2-w/o-d) & L + I + C & Hybrid     & 7.12m & 6.22°  & 0.45m  & 0.80°  & 3.27m & 5.92°  & 0.10m & 0.89° \\
\bottomrule
\end{tabular}
}
\end{table*}

\begin{figure}[t!]
    \centering
    \includegraphics[width=\linewidth]{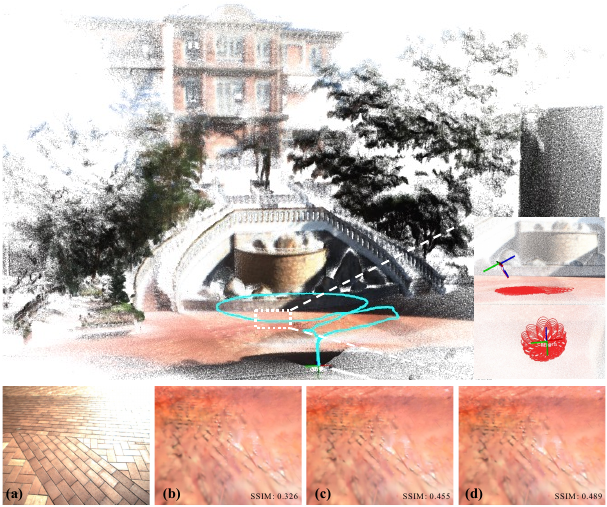}
    \caption{The trajectory and colored LiDAR point cloud map output by the continuous-time odometry efficiently fused with the Gaussian map. Even undergoing severe degradation when the solid-state LiDAR faces the plain ground, the odometry performs pose estimation accurately with minor start-to-end drift error, benefiting from the photometric constraints provided by the Gaussian map. (a)-(d) are the raw image of the current view and rendered images from poses predicted by the IMU, refined by Gaussian-map-based optimization, and further optimized through LiDAR-Inertial-Camera factor optimization, respectively. 
     Color discrepancies between the ground-truth and rendered images reflect variable illumination and varying camera exposure, highlighting the importance of the D-SSIM loss in Eq.\eqref{eq:tracking_loss}.
    }
    \label{fig:r3live}
\end{figure}

\begin{table}[t]
\centering
\caption{Localization Evaluation: The RMSE (m) of APE results on large-scale outdoor sequences in MCD dataset.}
\label{tab:loc_performance_tuhh}
\resizebox{\linewidth}{!}{
\begin{tabular}{lccc}
\toprule
            & tuhh\_day\_02 & tuhh\_day\_03 & tuhh\_day\_04 \\ \midrule
MonoGS~\cite{matsuki2024gaussian}      & 29.53         & 42.54         & 25.4          \\
Co-SLAM~\cite{wang2023co}     & 47.26         & $\times$             & $\times$             \\
SplaTAM~\cite{keetha2024splatam}     & $\times$             & 34.49         & $\times$             \\ 
MM3DGS-SLAM~\cite{sun2024mm3dgs} & 17.53         & 20.14         & 26.33         \\  
PIN-SLAM~\cite{pan2024pin}    & 0.16          & 0.11          & 0.38          \\ 
DBA-Fusion~\cite{zhou2024dba}  & 1.86          & 1.25          & 2.71          \\  
LVI-SAM~\cite{shan2021lvi}     & 0.14             & 0.13             & 0.18             \\
R3LIVE~\cite{lin2022r}  & 0.10          & 0.13          & 0.15          \\
FAST-LIVO2~\cite{zheng2024fast}  & \textbf{0.08}          & 0.11          & 0.13          \\
Gaussian-LIC~\cite{lang2025gaussian}  & \textbf{0.08}          & 0.09          & 0.10          \\
Gaussian-LIC2 (c1)      & \textbf{0.08}          & \textbf{0.08}          & \textbf{0.09}          \\
Gaussian-LIC2 (c2)      & \textbf{0.08}          & 0.09          & \textbf{0.09}          \\ \bottomrule
\end{tabular}
}
\end{table}

\subsubsection{Challenging Degenerate Sequences}
\label{sec:localize_in_degeneration}

Sequences \textit{degenerate\_seq\_00}, \textit{degenerate\_seq\_01}, \textit{LiDAR\_Degenerate}, and \textit{Visual\_Challenge}, provided in the R3LIVE and FAST-LIVO datasets, exhibit severe degradation, such as the
solid-state LiDAR with small FoV facing the ground or walls,
and the camera facing textureless surfaces or undergoing aggressive motions (\textit{Visual\_Challenge}). Tab.~\ref{tab:challenging_seq_comparison} reports the start-to-end drift (and rotation) errors. Methods that rely solely on the rendering loss from radiance field maps for localization, including MonoGS, Co-SLAM, SplaTAM, and MM3DGS-SLAM, achieve excellent localization accuracy in confined indoor scenarios with moderate view changes. However, they tend to fail in large-scale outdoor scenes with significant viewpoint variation, as the camera can easily enter regions where the map optimization has not yet converged. 
MonoGS is the only one that lacks an absolute scale, we align its estimated trajectory with that of Gaussian-LIC2 (c1) for obtaining the absolute scale. 
MM3DGS-SLAM, aided by IMU, completes
the sequence \textit{degenerate\_seq\_00}, but still fails on the rest. 
Based on a point-based NeRF presentation, PIN-SLAM optimizes the poses via point-to-model SDF loss~\cite{pan2024pin}, which is similar to the point-to-plane metric in ICP. It builds a purely geometric map and achieves robust pose optimization even in challenging lighting conditions, but suffers in LiDAR degradation scenarios where environmental structures fail to provide sufficient constraints for pose optimization.
Moreover, relying solely on the LiDAR data and constant velocity assumption for initial pose,
PIN-SLAM crashes on the sequence \textit{Visual\_Challenge} where aggressive motions happen. 
As a learning-based visual-inertial odometry system, DBA-Fusion exhibits greater robustness to LiDAR degradation, and its tight fusion with IMU data facilitates handling of aggressive motions. Nevertheless, excessive view changes in the sequence \textit{Visual\_Challenge} adversely affect the accuracy of the predicted dense optical flow thus significantly deteriorates the pose tracking in DBA-Fusion.

Among all the comparisons, LiDAR-Inertial-Camera fusion-based methods with each module meticulously designed showcase more robust and accurate performance. Rather than separately fusing LiDAR-inertial and visual-inertial data like LVI-SAM, R3LIVE, and FAST-LIVO2, our method maintains a single unified system that jointly and tightly integrates LiDAR-Inertial-Camera data within a continuous-time factor graph, achieving superior overall performance. Note that due to the lack of rigorous hardware synchronization on the sequences \textit{degenerate\_seq\_00-01}, the performance of FAST-LIVO2 is impacted. Fig.~\ref{fig:r3live} shows the trajectory and colored LiDAR map produced by Gaussian-LIC2 (c2), where the odometry successfully overcomes severe LiDAR degradation challenges and returns to the origin with minor drift. A set of subfigures in Fig.~\ref{fig:r3live} presents rendered images from three different viewpoints corresponding to poses predicted by the IMU, refined by Gaussian-map-based optimization (${}^W_C \widetilde{\mathbf{T}}$ in Eq.~\eqref{eq:map_optimize_pose}), and further refined through continuous-time LiDAR-Inertial-Camera factor graph optimization (Sec.~\ref{sec:lic_factorgraph}). The increasing similarity between the rendered and raw images visually demonstrates the progressive refinement of the estimated pose. Interestingly, Gaussian-LIC2 (c2) outperforms Gaussian-LIC2 (c1) on the \textit{Visual\_Challenge} sequence, which features textureless white walls. This improvement can be attributed to the photometric constraints derived from the Gaussian map, which enhance the robustness of Gaussian-LIC2 (c2) in low-texture environments compared to the optical-flow-based Gaussian-LIC2 (c1). Without depth supervision, Gaussian-LIC2 (c2-w/o-d) shows reduced accuracy. The depth regularization prevents the Gaussian map from overfitting to training views, which enables more robust novel view synthesis with fewer artifacts, as discussed in Sec.~\ref{sec:mapping}. This is critical since localization based on Gaussian maps inherently requires continuous novel view synthesis from varying perspectives. More results on other severe degenerate sequences from FAST-LIVO2~\cite{zheng2024fast} are available in the supplementary material (Sec.~\ref{sec:degenerated_fastlivo2}).

\subsubsection{Large-Scale Outdoor Sequences}
Tab.~\ref{tab:loc_performance_tuhh} displays the localization error across different methods in the large-scale scenes without sensor degradation. Rendering-based methods including MonoGS, Co-SLAM, SplaTAM, and MM3DGS-SLAM, exhibit reduced accuracy in pose estimation. Although MM3DGS-SLAM benefits from IMU-based pose initialization, its loosely-coupled fusion without accounting for IMU bias limits accuracy. In contrast, PIN-SLAM and DBA-Fusion perform better, with PIN-SLAM in particular approaching the performance of traditional multi-sensor-based methods. 
Attributed to the continuous-time trajectory representation, which effectively handles LiDAR distortion and efficiently fuses high-rate IMU data, our method achieves the best localization accuracy. In non-degenerate scenarios, visual information has a relatively minor impact on tracking accuracy, resulting in similar performance between Gaussian-LIC2 (c1) and Gaussian-LIC2 (c2).

\begin{figure*}[!htbp]
    \begin{minipage}{\linewidth}
        \makebox[0.16\linewidth]{\footnotesize GT-RGB}%
        \makebox[0.175\linewidth]{\footnotesize MonoGS}%
        \makebox[0.16\linewidth]{\footnotesize Co-SLAM}%
        \makebox[0.18\linewidth]{\footnotesize MM3DGS-SLAM}%
        \makebox[0.17\linewidth]{\footnotesize Gaussian-LIC}%
        \makebox[0.145\linewidth]{\footnotesize Gaussian-LIC2}%
    \end{minipage}
    \centering
    \includegraphics[width=\linewidth]{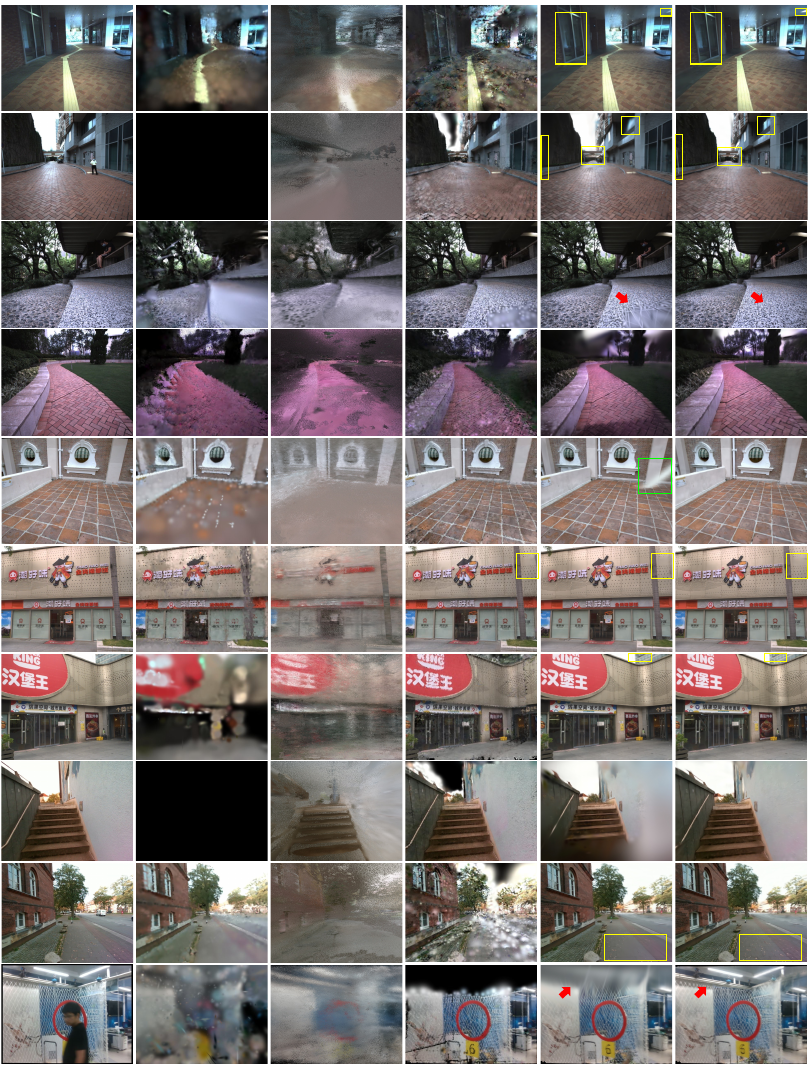}
    \caption{Qualitative results of RGB rendering (in-sequence novel view) on public datasets. Regions not observed by the LiDAR throughout the entire process are indicated by the red arrow, floaters are highlighted with green boxes, and key details are marked with yellow boxes. Gaussian-LIC2 consistently achieves sharper visual effects with fewer artifacts. 
     }
    \label{fig:in-seq-rgb}
\end{figure*}
\begin{figure*}[!htbp]
    \begin{minipage}{\linewidth}
        \makebox[0.16\linewidth]{\footnotesize GT-RGB}%
        \makebox[0.175\linewidth]{\footnotesize LiDAR-Depth}%
        \makebox[0.16\linewidth]{\footnotesize Co-SLAM}%
        \makebox[0.18\linewidth]{\footnotesize MM3DGS-SLAM}%
        \makebox[0.17\linewidth]{\footnotesize Gaussian-LIC}%
        \makebox[0.145\linewidth]{\footnotesize Gaussian-LIC2}%
    \end{minipage}
    \centering
    \includegraphics[width=\textwidth]{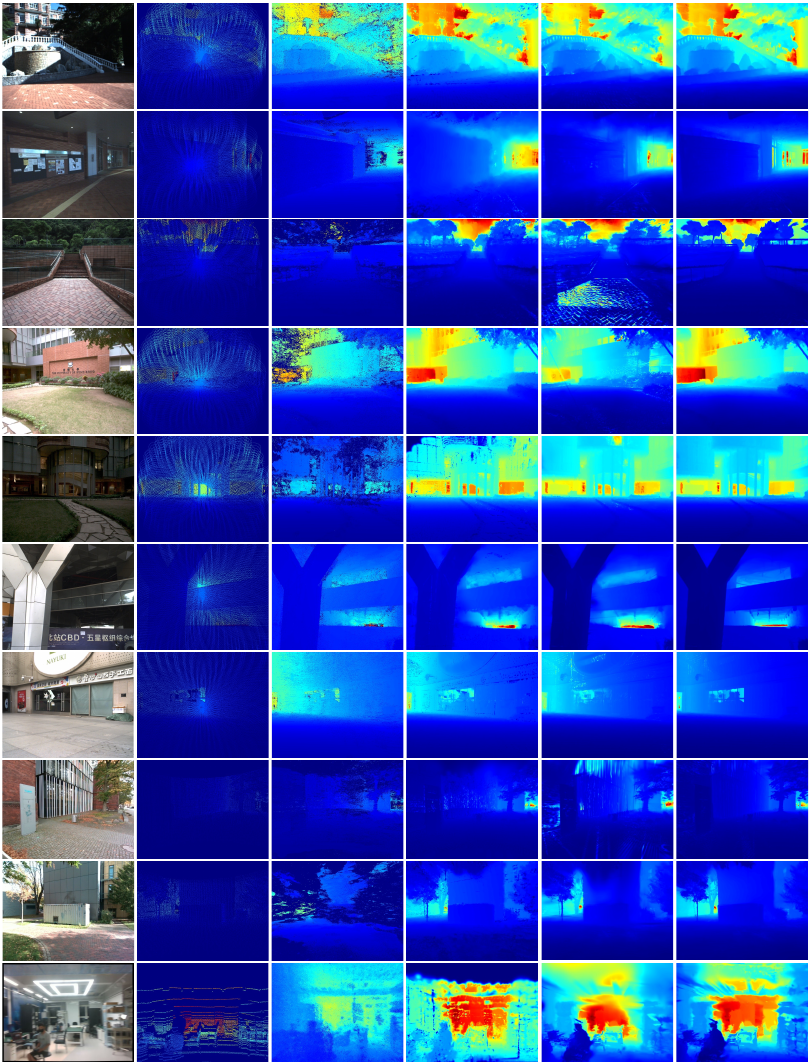}
     \caption{Qualitative results of depth rendering (in-sequence novel view) on public datasets. Sparse LiDAR depth at the viewpoints is also shown in the second column. Note that the last row shows the rendering results on the M2DGR dataset, which is impacted by the rolling shutter effect. Gaussian-LIC2 consistently achieves sharper visual effects with clearer geometric structures.
     }
    \label{fig:in-seq-dep}
\end{figure*}

\begin{table*}[t]
\centering
\caption{Evaluation of RGB rendering (in-sequence novel view) on public datasets.}
\label{tab:rgb_rendering}
\resizebox{\linewidth}{!}{
\begin{tabular}{cccccc} \toprule
\multirow{2}{*}{Sequence} &
  \multicolumn{5}{c}{Rendering Performance (PSNR$\uparrow$ SSIM$\uparrow$ LPIPS$\downarrow$)} \\ \cmidrule(lr){2-6} 
 & MonoGS & Co-SLAM & MM3DGS-SLAM & Gaussian-LIC & Gaussian-LIC2 \\ \midrule
 \cmidrule(lr){1-6}
\multicolumn{6}{l}{\cellcolor[HTML]{EEEEEE}{\textbf{R3LIVE (Avia)}}}                      \\
degenerate\_seq\_00 &
  14.59 $|$ 0.460 $|$ 0.624 &
  14.42 $|$ 0.286 $|$ 0.810 &
  19.35 $|$ 0.698 $|$ 0.212 &
  21.94 $|$ 0.777 $|$ 0.171 &
  \textbf{21.99} $|$ \textbf{0.782} $|$ \textbf{0.162} \\ 
degenerate\_seq\_01 & 14.66 $|$ 0.438 $|$ 0.692 & 14.88 $|$ 0.267 $|$ 0.822 & 18.57 $|$ 0.621 $|$ 0.277 & 22.80 $|$ 0.787 $|$ 0.169 & \textbf{22.94} $|$ \textbf{0.790} $|$ \textbf{0.159} \\
hku\_campus\_00     & 15.41 $|$ 0.464 $|$ 0.675 & 15.50 $|$ 0.302 $|$ 0.794 & 18.70 $|$ 0.621 $|$ 0.319 & \textbf{24.38} $|$ \textbf{0.787} $|$ 0.162 & 24.34 $|$ \textbf{0.787} $|$ \textbf{0.156} \\
hku\_campus\_01     & 8.39 $|$ 0.065 $|$ 0.873  & 12.92 $|$ 0.294 $|$ 0.796 & 16.97 $|$ 0.593 $|$ 0.271 & \textbf{20.55} $|$ 0.671 $|$ 0.275 & 20.52 $|$ \textbf{0.674} $|$ \textbf{0.236} \\
hku\_park\_00       & 13.24 $|$ 0.319 $|$ 0.733              & 14.02 $|$ 0.218 $|$ 0.790 & 14.98 $|$ 0.363 $|$ 0.396 & 17.09 $|$ 0.466 $|$ 0.354 & \textbf{17.17} $|$ \textbf{0.469} $|$ \textbf{0.340} \\
hku\_park\_01       & 12.22 $|$ 0.289 $|$ 0.790 & 13.65 $|$ 0.207 $|$ 0.815 & 15.79 $|$ 0.392 $|$ 0.409 & 19.45 $|$ \textbf{0.529} $|$ 0.340 & \textbf{19.46} $|$ \textbf{0.529} $|$ \textbf{0.325} \\ \midrule
\cmidrule(lr){1-6}
\multicolumn{6}{l}{\cellcolor[HTML]{EEEEEE}{\textbf{FAST-LIVO (Avia)}}}                      \\
LiDAR\_Degenerate &
  25.44 $|$ 0.779 $|$ 0.645 &
  23.59 $|$ 0.622 $|$ 0.496 &
  28.58 $|$ 0.828 $|$ 0.157 &
  30.09 $|$ 0.829 $|$ 0.155 &
  \textbf{30.36} $|$ \textbf{0.831} $|$ \textbf{0.144} \\
Visual\_Challenge   & 15.36 $|$ 0.563 $|$ 0.674 & 15.96 $|$ 0.365 $|$ 0.711 & 17.62 $|$ 0.712 $|$ 0.240 & 23.26 $|$ 0.821 $|$ 0.162 & \textbf{23.44} $|$ \textbf{0.822} $|$ \textbf{0.157} \\
hku1                & 13.44 $|$ 0.385 $|$ 0.787 & 15.98 $|$ 0.269 $|$ 0.738 & 21.76 $|$ 0.713 $|$ 0.162 & \textbf{23.82} $|$ 0.757 $|$ 0.153 & 23.74 $|$ \textbf{0.758} $|$ \textbf{0.149} \\
hku2                & 21.24 $|$ 0.559 $|$ 0.691 & 20.07 $|$ 0.398 $|$ 0.631 & 26.29 $|$ 0.754 $|$ 0.156 &\textbf{29.11} $|$ \textbf{0.798} $|$ 0.153 & 29.08 $|$ \textbf{0.798} $|$ \textbf{0.148} \\
Retail\_Street      & 18.74 $|$ 0.574 $|$ 0.537 & 17.32 $|$ 0.375 $|$ 0.669 & 21.55 $|$ 0.694 $|$ 0.162 & 24.15 $|$ 0.770 $|$ 0.128 & \textbf{24.37} $|$ \textbf{0.775} $|$ \textbf{0.121} \\
CBD\_Building\_01   & 17.11 $|$ 0.630 $|$ 0.640 & 18.16 $|$ 0.508 $|$ 0.644 & 22.13 $|$ 0.812 $|$ 0.126 & 25.16 $|$ \textbf{0.851} $|$ 0.104 & \textbf{25.20} $|$ \textbf{0.851} $|$ \textbf{0.103} \\ \midrule
\cmidrule(lr){1-6}
\multicolumn{6}{l}{\cellcolor[HTML]{EEEEEE}{\textbf{MCD (OS1-64)}}}                      \\
tuhh\_day\_02 &
  8.72 $|$ 0.158 $|$ 0.893 &
  12.65 $|$ 0.295 $|$ 0.766 &
  11.15 $|$ 0.478 $|$ 0.350 &
  19.99 $|$ 0.621 $|$ 0.312 &
  \textbf{20.35} $|$ \textbf{0.626} $|$ \textbf{0.262} \\
tuhh\_day\_03       & 8.09 $|$ 0.136 $|$ 0.896  & 12.48 $|$ 0.407 $|$ 0.684 & 11.36 $|$ 0.542 $|$ 0.301 & 21.09 $|$ 0.666 $|$ 0.273 & \textbf{21.38} $|$ \textbf{0.672} $|$ \textbf{0.229} \\
tuhh\_day\_04       & 11.73 $|$ 0.250 $|$ 0.805  & 13.02 $|$ 0.413 $|$ 0.626 & 13.86 $|$ 0.384 $|$ 0.398 & \textbf{19.27} $|$ \textbf{0.528} $|$ 0.329 & \textbf{19.27} $|$ \textbf{0.528} $|$ \textbf{0.310} \\ \midrule
\cmidrule(lr){1-6}
\multicolumn{6}{l}{\cellcolor[HTML]{EEEEEE}{\textbf{M2DGR (VLP-32C)}}}                      \\
room\_01 &
  14.92 $|$ 0.573 $|$ 0.643 &
  15.01 $|$ 0.317 $|$ 0.880 &
  12.27 $|$ 0.554 $|$ 0.458 &
  17.04 $|$ 0.697 $|$ 0.365 &
  \textbf{17.45} $|$ \textbf{0.721} $|$ \textbf{0.282} \\
room\_02            & 13.57 $|$ 0.515 $|$ 0.697 & 14.94 $|$ 0.318 $|$ 0.877 & 11.44 $|$ 0.565 $|$ 0.465 & 17.32 $|$ 0.705 $|$ 0.375 & \textbf{17.72} $|$ \textbf{0.725} $|$ \textbf{0.291} \\
room\_03            & 16.15 $|$ 0.627 $|$ 0.587 & 14.88 $|$ 0.370 $|$ 0.841 & 12.50 $|$ 0.573 $|$ 0.475 & 17.19 $|$ 0.701 $|$ 0.401 & \textbf{17.38} $|$ \textbf{0.709} $|$ \textbf{0.334} \\
\bottomrule
\end{tabular}
}
\end{table*}

\subsection{Experiment-2: Evaluation of Mapping}
\label{sec:mapping}

\subsubsection{Evaluation Protocols}
We assess the performance of photo-realistic mapping by evaluating the quality of rendered images generated from the Gaussian map. To this end, we adopt several widely used metrics, including Peak Signal-to-Noise Ratio (PSNR), Structural Similarity Index (SSIM), and Learned Perceptual Image Patch Similarity (LPIPS)~\cite{kerbl20233d}. Consistent with MonoGS and SplaTAM, we use AlexNet~\cite{krizhevsky2012imagenet} as the backbone network for LPIPS evaluation. When ground-truth depth maps are available, we also evaluate geometric accuracy using Depth-L1 error computed over valid depth regions. Additionally, to better compare the geometric quality, we further assess the commonly used surface reconstruction metrics, namely accuracy, completeness, Chamfer-L1 distance, and F-score. It should be noted that we adopt Gaussian-LIC2 (c1) in the mapping evaluation and all the following experiments. 

Many existing methods~\cite{xie2024gs, zhao2025lvi, hong2025gs} report rendering performance at their respective training views. However, such evaluations may be biased due to overfitting and fail to reliably reflect photo-realistic mapping quality at novel views. To ensure a comprehensive and fair evaluation, we assess rendering quality on both in-sequence and out-of-sequence novel views, explicitly excluding training views.

For fair evaluations, we provide accurate poses for all methods to isolate mapping performance from pose estimation errors, and all methods are trained using the same keyframe set  (Sec.~\ref{sec:grouping}), with non-keyframes reserved for in-sequence novel view rendering. 
Specifically, for our self-collected dataset where ground-truth poses are available, all methods use ground-truth poses, while on public datasets lacking ground-truth, both our method and the baselines are provided with our estimated poses for evaluation. 

\subsubsection{In-Sequence Novel View Synthesis}
Tab.~\ref{tab:rgb_rendering} and Tab.~\ref{tab:rendering_oos} show quantitative rendering results of in-sequence novel views across both public and self-collected datasets. Fig.~\ref{fig:in-seq-rgb} and Fig.~\ref{fig:in-seq-dep} present the novel view renderings of RGB and depth in sequence.
The RGB-only method MonoGS initializes Gaussians at a preset constant depth with random noise and subsequently inserts new Gaussians based on the rendered depth statistics.
While the optimization of the inserted Gaussians may gradually converge in small-scale indoor scenarios, it becomes much more challenging in larger scenes, where misplaced Gaussians accumulate, causing floaters and poor visual quality. 
Incorporating LiDAR geometric priors can substantially alleviate this problem. For example, Co-SLAM leverages the LiDAR depth for ray sampling and supervision of the neural implicit map optimization, rendering relatively clear structures. It is capable of rendering beyond the LiDAR FoV based on the optimized neural representation, as shown in Fig.~\ref{fig:in-seq-rgb} and Fig.~\ref{fig:in-seq-dep}. However, the renderings of Co-SLAM appear noisy with severe artifacts, even when guided by completed depth maps.
In contrast, the LiDAR-incorporated 3DGS-based methods MM3DGS-SLAM and Gaussian-LIC exhibit better rendering performance. They reliably initialize Gaussians from precise LiDAR points. Nonetheless, MM3DGS-SLAM represents the scene as isotropic and view-independent Gaussians to accelerate, but unfortunately still fails to achieve real-time performance while sacrificing visual quality. Note that MM3DGS-SLAM uses around three times as many Gaussians as our method to model the scene. 
Gaussian-LIC underutilizes precise LiDAR depth and ignores to optimize the geometric quality of the Gaussian map, leading to poor-quality rendered depth. In addition, both methods struggle to accurately reconstruct areas unobserved by the LiDAR.

Gaussian-LIC2 attains the best in-sequence novel view rendering performance, both quantitatively and qualitatively. The system renders sharper RGB images with fewer artifacts at novel views, beyond only overfitting to the training views. It successfully reconstructs regions never scanned by the LiDAR, such as the ceiling beyond the reach of the VLP-32C, as demonstrated in the last row of Fig.~\ref{fig:in-seq-rgb}. Moreover, despite relying on sparse LiDAR depth, our method is able to produce high-quality depth maps across different LiDAR modalities. It is worth noting that severe rolling shutter distortion in the images of the M2DGR dataset affects the quality of the rendered depth, as illustrated in the last row of Fig.~\ref{fig:in-seq-dep}.

\subsubsection{Out-of-Sequence Novel View Synthesis}
\label{sec:oos}
\begin{figure}[!t]
    \begin{minipage}{\linewidth}
        \makebox[0.25\linewidth]{\scriptsize GT}%
        \makebox[0.25\linewidth]{\scriptsize MM3DGS-SLAM}%
        \makebox[0.25\linewidth]{\scriptsize Gaussian-LIC}%
        \makebox[0.25\linewidth]{\scriptsize Gaussian-LIC2}%
    \end{minipage}
    \centering
    \includegraphics[width=\linewidth]{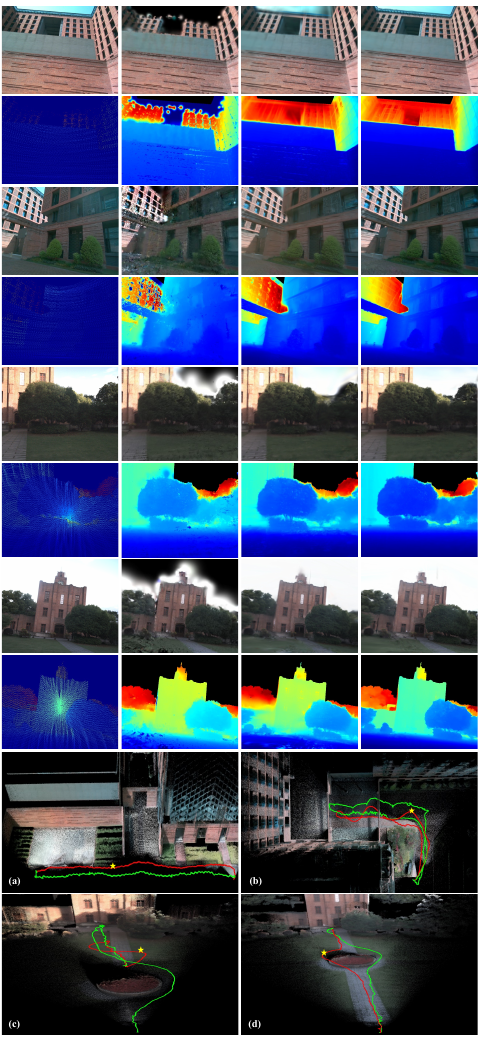}
    \caption{Qualitative results of RGB and depth rendering (out-of-sequence novel view) on our self-collected dataset. (a-d) The green path represents the trajectory for collecting training views, the red path shows the out-of-sequence trajectory for evaluation, and the yellow stars indicate the selected out-of-sequence novel views. The sky regions in rendered depth maps are masked in black with the segmentation model. }
    \label{fig:out-of-seq}
\end{figure}

We further evaluate out-of-sequence novel view rendering on our self-collected dataset, which presents a greater challenge than in-sequence rendering. For this assessment, views are sampled at 10 Hz along the out-of-sequence trajectory. As shown in Tab.~\ref{tab:rendering_oos}, our method achieves the best results. Fig.~\ref{fig:out-of-seq} illustrates the camera trajectories and the corresponding novel view rendering results along the out-of-sequence path. Compared to Gaussian-LIC, the Gaussian map of Gaussian-LIC2 is more effectively regularized and optimized, leading to sharper and higher-fidelity novel view synthesis with fewer artifacts.

\begin{table*}[t]
\centering
\caption{Evaluation of RGB and depth rendering (in-sequence $\&$ out-of-sequence novel view) on our datasets. Depth-L1 (m).}
\label{tab:rendering_oos}
\resizebox{\linewidth}{!}{
\begin{tabular}{ccccccccccc}
\toprule
\multirow{2}{*}{LiDAR} & \multirow{2}{*}{Sequence}  & \multirow{2}{*}{Method}  & \multicolumn{4}{c}{In-Sequence Novel View} & \multicolumn{4}{c}{Out-of-Sequence Novel View} \\
\cmidrule(lr){4-7} \cmidrule(lr){8-11}
& & & PSNR$\uparrow$ & SSIM$\uparrow$ & LPIPS$\downarrow$ & Depth-L1$\downarrow$ & PSNR$\uparrow$ & SSIM$\uparrow$ & LPIPS$\downarrow$ & Depth-L1$\downarrow$ \\
\midrule
\multirow{15}{*}{ Mid-360 } & \multirow{3}{*}{ Liberal\_Arts\_Group\_01} & MM3DGS-SLAM & 22.89 & 0.721 & 0.298 & 0.65 & 21.05 & 0.653 & 0.360 & 0.65 \\
& & Gaussian-LIC & 25.12 & 0.759 & 0.297 & 0.92 & 22.08 & 0.670 & 0.350 & 0.81 \\
& & Gaussian-LIC2 & \textbf{25.35} & \textbf{0.761} & \textbf{0.278} & \textbf{0.29} & \textbf{22.10} & \textbf{0.671} & \textbf{0.332} & \textbf{0.30} \\ \cmidrule(lr){2-11}
& \multirow{3}{*}{Liberal\_Arts\_Group\_02} & MM3DGS-SLAM & 17.96 & 0.574 & 0.271 & 0.67 & 16.98 & 0.444 & 0.351 & 0.72 \\
& & Gaussian-LIC & 23.02 & 0.703 & 0.210 & 0.66 & 19.95 & 0.520 & 0.275 & 0.71 \\
& & Gaussian-LIC2 & \textbf{23.42} & \textbf{0.713} & \textbf{0.186} & \textbf{0.30} & \textbf{20.05} & \textbf{0.528} & \textbf{0.256} & \textbf{0.37} \\ \cmidrule(lr){2-11}
& \multirow{3}{*}{Liberal\_Arts\_Group\_03} & MM3DGS-SLAM & 17.74 & 0.493 & 0.385 & 0.96 & 13.87 & 0.347 & 0.478 & 1.22 \\
& & Gaussian-LIC & 22.59 & 0.656 & 0.236 & 0.89 & \textbf{18.28} & 0.458 & 0.351 & 1.25 \\
& & Gaussian-LIC2 & \textbf{22.79} & \textbf{0.662} & \textbf{0.222} & \textbf{0.44} & \textbf{18.28} & \textbf{0.459} & \textbf{0.345} & \textbf{0.67} \\ \cmidrule(lr){2-11}
& \multirow{3}{*}{Medical\_Building\_01} & MM3DGS-SLAM & 16.45 & 0.642 & 0.266 & 1.24 & 16.43 & 0.574 & 0.321 & 1.34 \\
& & Gaussian-LIC & 22.77 & 0.737 & 0.199 & 1.28 & 19.40 & 0.632 & 0.268 & 1.64 \\
& & Gaussian-LIC2 & \textbf{22.89} & \textbf{0.741} & \textbf{0.188} & \textbf{0.62} & \textbf{19.54} & \textbf{0.633} & \textbf{0.257} & \textbf{0.84} \\ \cmidrule(lr){2-11}
& \multirow{3}{*}{Medical\_Building\_02} & MM3DGS-SLAM & 16.29 & 0.685 & 0.217 & 0.91 & 19.62 & 0.679 & 0.222 & 0.75 \\
& & Gaussian-LIC & \textbf{23.15} & 0.758 & 0.177 & 1.53 & 21.33 & 0.685 & 0.216 & 1.35 \\
& & Gaussian-LIC2 & 23.11 & \textbf{0.760} & \textbf{0.174} & \textbf{0.54} & \textbf{21.57} & \textbf{0.691} & \textbf{0.213} & \textbf{0.53} \\ \midrule
\multirow{15}{*}{ Avia } & \multirow{3}{*}{Lecture\_Hall} & MM3DGS-SLAM & 19.54 & 0.671 & 0.198 & 0.74 & 18.73 & 0.624 & 0.236 & 0.75 \\
& & Gaussian-LIC & 21.60 & 0.710 & 0.178 & 0.68 & 20.32 & 0.669 & 0.207 & 0.73 \\
& & Gaussian-LIC2 & \textbf{21.78} & \textbf{0.719} & \textbf{0.167} & \textbf{0.39} & \textbf{20.40} & \textbf{0.672} & \textbf{0.205} & \textbf{0.45} \\ \cmidrule(lr){2-11}
& \multirow{3}{*}{Robot\_Center} & MM3DGS-SLAM & 18.83 & 0.631 & 0.190 & 0.59 & 16.63 & 0.558 & 0.260 & 0.47 \\
& & Gaussian-LIC & 22.89 & 0.693 & 0.206 & 0.92 & \textbf{18.81} & 0.600 & 0.264 & 0.79 \\
& & Gaussian-LIC2 & \textbf{22.95} & \textbf{0.699} & \textbf{0.188} & \textbf{0.29} & 18.80 & \textbf{0.602} & \textbf{0.257} & \textbf{0.25} \\ \cmidrule(lr){2-11}
& \multirow{3}{*}{Bell\_Tower\_01} & MM3DGS-SLAM & 18.13 & 0.600 & 0.300 & 0.67 & 16.40 & 0.592 & 0.299 & 0.74 \\
& & Gaussian-LIC & 25.38 & 0.717 & 0.282 & 0.92 & \textbf{22.87} & \textbf{0.676} & 0.324 & 0.91 \\
& & Gaussian-LIC2 & \textbf{25.53} & \textbf{0.720} & \textbf{0.256} & \textbf{0.27} & \textbf{22.87} & \textbf{0.676} & \textbf{0.298} & \textbf{0.35} \\ \cmidrule(lr){2-11}
& \multirow{3}{*}{Bell\_Tower\_02} & MM3DGS-SLAM & 16.49 & 0.661 & 0.278 & 0.87 & 17.01 & 0.620 & 0.320 & 1.14 \\
& & Gaussian-LIC & 26.42 & 0.751 & 0.285 & 1.18 & 23.68 & \textbf{0.716} & 0.321 & 1.77 \\
& & Gaussian-LIC2 & \textbf{26.85} & \textbf{0.756} & \textbf{0.242} & \textbf{0.29} & \textbf{24.17} & 0.708 & \textbf{0.276} & \textbf{0.41} \\ \cmidrule(lr){2-11}
& \multirow{3}{*}{Bell\_Tower\_03} & MM3DGS-SLAM & 16.61 & 0.651 & 0.256 & 0.90 & 14.08 & 0.603 & 0.294 & 1.02 \\
& & Gaussian-LIC & 27.29 & 0.748 & 0.296 & 1.47 & 25.02 & 0.735 & 0.278 & 1.68 \\
& & Gaussian-LIC2 & \textbf{27.63} & \textbf{0.756} & \textbf{0.244} & \textbf{0.33} & \textbf{25.43} & \textbf{0.738} & \textbf{0.238} & \textbf{0.42} \\
\bottomrule
\end{tabular}
}
\end{table*}

\begin{table}[t]
\centering
\caption{Evaluation of surface reconstruction on self-collected dataset. Metrics in meters: accuracy (Acc.), completeness (Comp.), Chamfer-L1 distance (C-L1). F-score in $\%$ is calculated with a 0.2m error threshold. The results are averaged separately over the Mid-360 dataset and the Avia dataset. RT means real time. Best results are in bold and second-best results are underlined. Gaussian-LIC2* means the modified version in Sec.~\ref{sec:mesh} for the specific application of rapid mesh production.}
\label{tab:mid360_livox_geo}
\resizebox{\linewidth}{!}{%
\begin{tabular}{llcccccc}
\toprule
LiDAR & Method & Acc.$\downarrow$ & Comp.$\downarrow$ & C-L1$\downarrow$ & F-score$\uparrow$ & PSNR$\uparrow$ & RT \\
\midrule
\multirow{5}{*}{Mid-360}
& Co-SLAM        & 0.197 & 0.461 & 0.333 & 49.81 & 16.15 & $\times$ \\
& MM3DGS-SLAM    & 0.181 & 0.402 & 0.292 & 58.32 & 18.27 & $\times$ \\
& Gaussian-LIC   & 0.189 & 0.424 & 0.326 & 51.49 & \underline{23.33} & $\checkmark$ \\
& Gaussian-LIC2  & \underline{0.133} & \underline{0.218} & \underline{0.186} & \underline{70.77} & \textbf{23.51} & $\checkmark$ \\
& Gaussian-LIC2* & \textbf{0.125} & \textbf{0.178} & \textbf{0.162} & \textbf{75.34} & 21.95 & $\checkmark$ \\
\midrule
\multirow{5}{*}{Avia}
& Co-SLAM        & 0.226 & 0.490 & 0.351 & 46.42 & 13.55 & $\times$ \\
& MM3DGS-SLAM    & 0.152 & 0.377 & 0.264 & 61.21 & 17.92 & $\times$ \\
& Gaussian-LIC   & 0.149 & 0.442 & 0.295 & 59.21 & \underline{24.72} & $\checkmark$ \\
& Gaussian-LIC2  & \underline{0.107} & \underline{0.189} & \underline{0.147} & \underline{75.68} & \textbf{24.95} & $\checkmark$ \\
& Gaussian-LIC2* & \textbf{0.081} & \textbf{0.153} & \textbf{0.118} & \textbf{83.72} & 23.99 & $\checkmark$ \\
\bottomrule
\end{tabular}%
}
\end{table}

\subsubsection{Surface Reconstruction}
Tab.~\ref{tab:mid360_livox_geo} reports the geometric quality quantitatively on our self-collected dataset, where the denoised ground-truth point cloud is provided. The metrics are calculated from the ground-truth and the predicted mesh (Sec.~\ref{sec:mesh}). Note that we compute F-score with a 0.2m error threshold. As can be seen, with visual quality and real-time
performance as prerequisites, Gaussian-LIC2 achieves more accurate and more complete reconstruction of the environment than the compared methods. Furthermore, as in the application described in Sec.~\ref{sec:mesh}, by sacrificing a small amount of visual quality, Gaussian-LIC2 can achieve higher geometric quality, which can be specifically utilized for rapid mesh generation.

\subsection{Experiment-3: Offline Gaussian Mapping}
As a SLAM system, our method incrementally estimates poses and constructs a photo-realistic map from sequential sensor data, enabling real-time perception for robotic applications. In contrast, offline methods disregard incremental or online processing capabilities and typically perform heavy, computationally intensive optimization only after all sensor data has been collected. Existing offline 3DGS mapping methods typically fall into two paradigms, namely per-scene optimization and generalizable feed-forward models.

\subsubsection{Per-Scene Optimization}
We first compare our method to the state-of-the-art LiDAR-based 3DGS framework, LetsGo~\cite{cui2024letsgo}, on the 105-second \textit{Lecture\_Hall} sequence. To ensure a fair comparison, LetsGo is provided with the same sparse LiDAR depth maps, our estimated poses, and the same set of training views, trained for 30,000 iterations.
Tab.~\ref{tab:letsgo} presents the results, reporting the total time from data acquisition start to reconstruction completion.
Our method achieves comparable performance while completing the reconstruction within the duration of data acquisition, whereas LetsGo requires time-consuming optimization after receiving the full dataset.
Furthermore, our method significantly outperforms LetsGo after an additional 10,000 iterations of refinement (requiring only 24 seconds). This demonstrates that our approach not only supports real-time reconstruction but can also achieve higher accuracy when post-acquisition optimization is permitted.

\begin{table}[t]
\centering
\caption{Comparison with LiDAR-based offline Gaussian mapping method on the sequence
\textit{Lecture\_Hall} (span: 105 seconds). The total time consumption reported here is the duration of data acquisition time $+$ reconstruction time.}
\label{tab:letsgo}
\resizebox{\linewidth}{!}{
\begin{tabular}{lccccc}
\toprule
\textbf{} & PSNR$\uparrow$ & SSIM$\uparrow$ & LPIPS$\downarrow$ & Depth-L1 (m)$\downarrow$ & Time (s)$\downarrow$ \\
\midrule
LetsGo~\cite{cui2024letsgo}        & 21.81     & 0.708     & 0.170     & \textbf{0.37}     & 105 + 347    \\
Gaussian-LIC2          & 21.78 & 0.719 & 0.167 & 0.39  & \textbf{105}       \\
Gaussian-LIC2-W-Refine   & \textbf{22.28} & \textbf{0.730} & \textbf{0.148} & \textbf{0.37}  & 105 + 24  \\
\bottomrule
\end{tabular}}
\end{table}

\begin{figure}[t!]
    \centering
    \includegraphics[width=\linewidth]{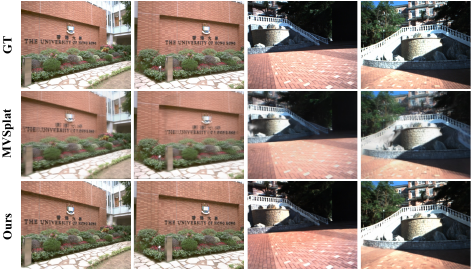}
    \caption{Comparison with the feed-forward method, MVSplat~\cite{chen2024mvsplat}. Image renderings from both MVSplat and Gaussian-LIC2 are shown.}
    \label{fig:feed-forward}
\end{figure}

\subsubsection{Feed-Forward Model}
We also compare our method with the state-of-the-art feed-forward Gaussian mapping approach, MVSplat~\cite{chen2024mvsplat}, which predicts a Gaussian map from a small set of posed images (two by default). For achieving good performance with MVSplat, we carefully select a pair of images that provides sufficient parallax and maximal scene coverage, using poses estimated by our method. Fig.~\ref{fig:feed-forward} shows the rendering results on the \textit{degenerate\_seq\_00} and \textit{hku1} sequences. While MVSplat is capable of predicting pixel-aligned Gaussians within seconds, it often produces blurry renderings and suffers from parallax sensitivity. On the contrary, our method consistently generates sharp, photo-realistic images.

\subsection{Runtime Analysis}
\begin{table}[t]
\centering
\caption{Runtime Analysis (Time unit: s). Fwd.: forward time; Bwd.: backward time; Adam: optimizer update time; Total: total mapping time; Dur.: rosbag duration; Len.: trajectory length (m).}
\label{tab:runtime}
\resizebox{\linewidth}{!}{
\begin{tabular}{ccccccc}
\toprule
& \textbf{Fwd.} & \textbf{Bwd.} & \textbf{Adam} & \textbf{Total} & \textbf{Dur.} & \textbf{Len.} \\ \midrule
\multicolumn{7}{l}{\cellcolor[HTML]{EEEEEE}{\textbf{R3LIVE}}} \\      
degenerate\_seq\_00            & 9   & 16  & 9  & 38  & 87  & 42  \\
degenerate\_seq\_01            & 11  & 15  & 8  & 38  & 86  & 61  \\
hku\_campus\_seq\_00           & 69  & 69  & 45 & 195 & 202 & 172 \\
hku\_campus\_seq\_01           & 154 & 73  & 48 & 285 & 304 & 337 \\
hku\_park\_00                  & 62  & 87  & 55 & 218 & 228 & 212 \\
hku\_park\_01                  & 138 & 112 & 73 & 340 & 362 & 354 \\ 
\cmidrule(lr){1-7}
\multicolumn{7}{l}{\cellcolor[HTML]{EEEEEE}{\textbf{FAST-LIVO}}} \\
LiDAR\_Degenerate              & 7   & 12  & 6  & 29  & 78  & 38  \\
Visual\_Challenge              & 18  & 29  & 14 & 71  & 162 & 78  \\
hku1                           & 20  & 28  & 15 & 71  & 128 & 64  \\
hku2                           & 13  & 21  & 11 & 51  & 105 & 59  \\
Retail\_Street                  & 15  & 26  & 15 & 64  & 135 & 66  \\
CBD\_Building\_01                & 12  & 19  & 10 & 48  & 119 & 33  \\
\cmidrule(lr){1-7}
\multicolumn{7}{l}{\cellcolor[HTML]{EEEEEE}{\textbf{MCD}}} \\
tuhh\_day\_02                  & 74  & 57  & 35 & 179 & 200 & 306 \\
tuhh\_day\_03                  & 70  & 56  & 34 & 173 & 200 & 276 \\
tuhh\_day\_04                  & 43  & 46  & 27 & 127 & 187 & 298 \\
\cmidrule(lr){1-7}
\multicolumn{7}{l}{\cellcolor[HTML]{EEEEEE}{\textbf{M2DGR}}} \\
room01                         & 9   & 17  & 11 & 40  & 75  & 27  \\
room02                         & 8   & 16  & 10 & 38  & 89  & 46  \\
room03                         & 21  & 43  & 28 & 100 & 195 & 71  \\
\cmidrule(lr){1-7}
\multicolumn{7}{l}{\cellcolor[HTML]{EEEEEE}{\textbf{Self-collected}}} \\
Liberal\_Arts\_Group\_01       & 8   & 9   & 3  & 24  & 101 & 42  \\
Liberal\_Arts\_Group\_02        & 23  & 35  & 16 & 88  & 225 & 59  \\
Liberal\_Arts\_Group\_03        & 39  & 50  & 25 & 132 & 285 & 102 \\
Medical\_Building\_01            & 9   & 10  & 4  & 28  & 113 & 53  \\
Medical\_Building\_02            & 10  & 12  & 4  & 32  & 120 & 38  \\
Lecture\_Hall                  & 9   & 15  & 6  & 36  & 105 & 45  \\
Robot\_Center                  & 18  & 25  & 13 & 63  & 136 & 53  \\
Bell\_Tower\_01                & 36  & 30  & 15 & 90  & 150 & 67  \\
Bell\_Tower\_02                & 13  & 18  & 9  & 45  & 103 & 39  \\
Bell\_Tower\_03                & 18  & 19  & 9  & 52  & 120 & 48  \\ 
\bottomrule
\end{tabular}
}
\end{table}

We evaluate the real-time performance across all datasets. Following the definition in Gaussian-LIC~\cite{lang2025gaussian}, a system is considered to be real-time capable if it completes processing within the duration of sensor data acquisition, without extensive post-processing.
We evaluate every module of Gaussian-LIC(c1) during runtime analysis. The average pose estimation frequency is 10 Hz, while depth completion inference requires only 10–20 ms per frame.
Therefore, we primarily focus on the time consumption in the mapping thread. 
As shown in Tab.~\ref{tab:runtime}, the most time-consuming components are primarily the forward and backward pass of the rasterizer as well as the Adam optimizer update. Thanks to carefully designed CUDA-related acceleration strategies in Sec.~\ref{sec:acc}, the total mapping time remains within the bag duration, showcasing the real-time capability of our method. We also report the runtime of other methods and CUDA ablation in the supplementary material (Sec.~\ref{sec:runtime_ablation}). The real-time demonstration video can be found on the website\footnote{\url{https://www.youtube.com/watch?v=SkPnpuCfh88}}.

\section{Applications}

\begin{figure}[t!]
    \centering
    \includegraphics[width=\linewidth]{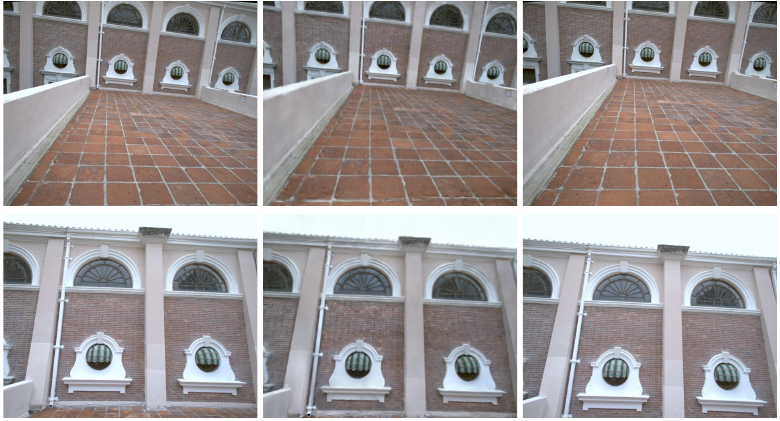}
    \caption{Application -- Video Frame Interpolation: The images in the middle are the interpolated frames at the intermediate timestamps of the left and right images.}
    \label{fig:video_interp}
\end{figure}
\begin{figure*}[t!]
    	\centering
    	\includegraphics[width=\textwidth]{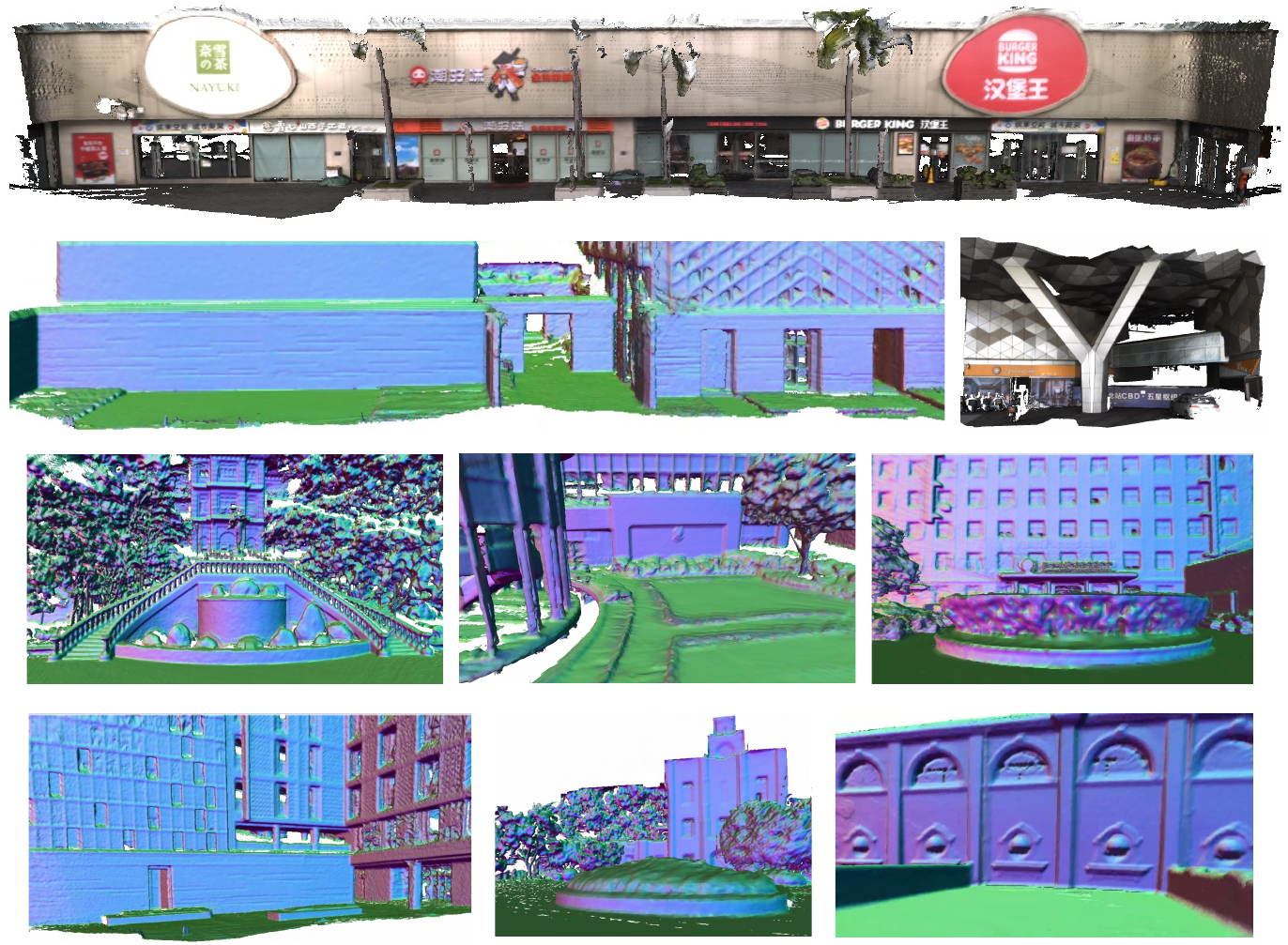}

    	\caption{Application -- Mesh Extraction: Textured and normal-colorized meshes generated from our reconstructed Gaussian map.
     }
    \label{fig:mesh}
\end{figure*}

\subsection{Video Frame Interpolation}
The continuous-time trajectory enables pose querying at any valid timestamp, while the Gaussian map allows rendering from arbitrary viewpoints. Interestingly, combining these capabilities facilitates spatiotemporal interpolation, which can be applied to video frame interpolation. As in Fig.~\ref{fig:video_interp}, after obtaining the continuous-time trajectory and Gaussian map from Gaussian-LIC2, we successfully double the frame rate of the sequence \textit{Visual\_Challenge} from 15 Hz to 30 Hz by rendering intermediate frames at the queried mid-time poses.

\subsection{Rapid 3D Mesh Extraction}
\label{sec:mesh}
Compared to previous LiDAR-Inertial-Camera 3DGS-based SLAM methods, Gaussian-LIC2 balances visual quality, geometric accuracy, and computational efficiency. In this application, we showcase how our method can be adapted for rapid meshing. We increase the depth regularization weight $\gamma$ (see Eq.\eqref{eq:depth_loss}) to be 2, and run our system in real time within the duration of data acquisition. Subsequently, based on the reconstructed Gaussian map, we render RGB images and depth maps from all viewpoints. These rendered depth maps are fused using TSDF fusion~\cite{curless1996volumetric} (with a voxel size of 0.05m) to reconstruct a 3D-consistent map. An accurate mesh is then extracted from the TSDF volume using the Marching Cubes algorithm~\cite{lorensen1998marching}. Fig.~\ref{fig:mesh} illustrates the resulting mesh, textured with RGB colors or colorized by surface normal directions.
\section{Conclusions and Future Work}
\label{sec:conclusion}

In this paper, We propose a novel LiDAR-Inertial-Camera Gaussian Splatting SLAM system that jointly considers visual quality, geometric accuracy, and real-time performance. Our method enables accurate pose estimation and photo-realistic map construction in real time, supporting high-quality RGB and depth rendering. Markedly, we incorporate a zero-shot depth completion model that fuses RGB and sparse LiDAR data to generate dense depth maps, which facilitate initialization of Gaussians in large-scale scenarios. The training of the Gaussian map is efficiently supervised by our curated sparse LiDAR depth and accelerated with meticulously designed CUDA-related strategies. Meanwhile, we explore jointly fusing the visual photometric constraints derived from the Gaussian map with the LiDAR-inertial data within the continuous-time framework, effectively overcoming the LiDAR degeneration. We also extend our system to support downstream tasks such as video interpolation and rapid mesh generation. Finally, we introduce a dedicated LiDAR-Inertial-Camera dataset for benchmarking photometric and geometric mapping in large-scale scenarios,  with ground-truth poses and depth maps, as well as out-of-sequence trajectories. Extensive experiments shows that our system outperforms existing methods in various aspects.

Our method enhances the real-time perception capabilities of mobile robotic systems in large-scale scenarios. However, several limitations remain for future work. (1) Map Compactness: We currently do not impose constraints on map size, which may result in large memory usage. It is worth investigating to reduce the number of Gaussians while maintaining the map quality.  (2) Geometric Accuracy: There remains a trade-off between visual quality and geometric accuracy. We aim to further improve geometric precision without sacrificing visual fidelity. (3) Applicability in Giga Scene: In extremely large-scale environments, our system may accumulate pose drift over time. We plan to introduce loop closures to mitigate this issue. (4) Integration with Foundation Models: A promising direction involves leveraging foundation models for feed-forward Gaussian prediction, combined with generative models, to improve novel view synthesis performance in out-of-sequence scenarios.

\begin{funding}
This work was supported by the National Natural Science Foundation
of China under Grant No. 62525309.
\end{funding}

\bibliography{main}

\appendix
\section{Supplementary Material}
\label{sec:supplementary}

\begin{figure}[t!]
    \centering
    \includegraphics[width=\linewidth]{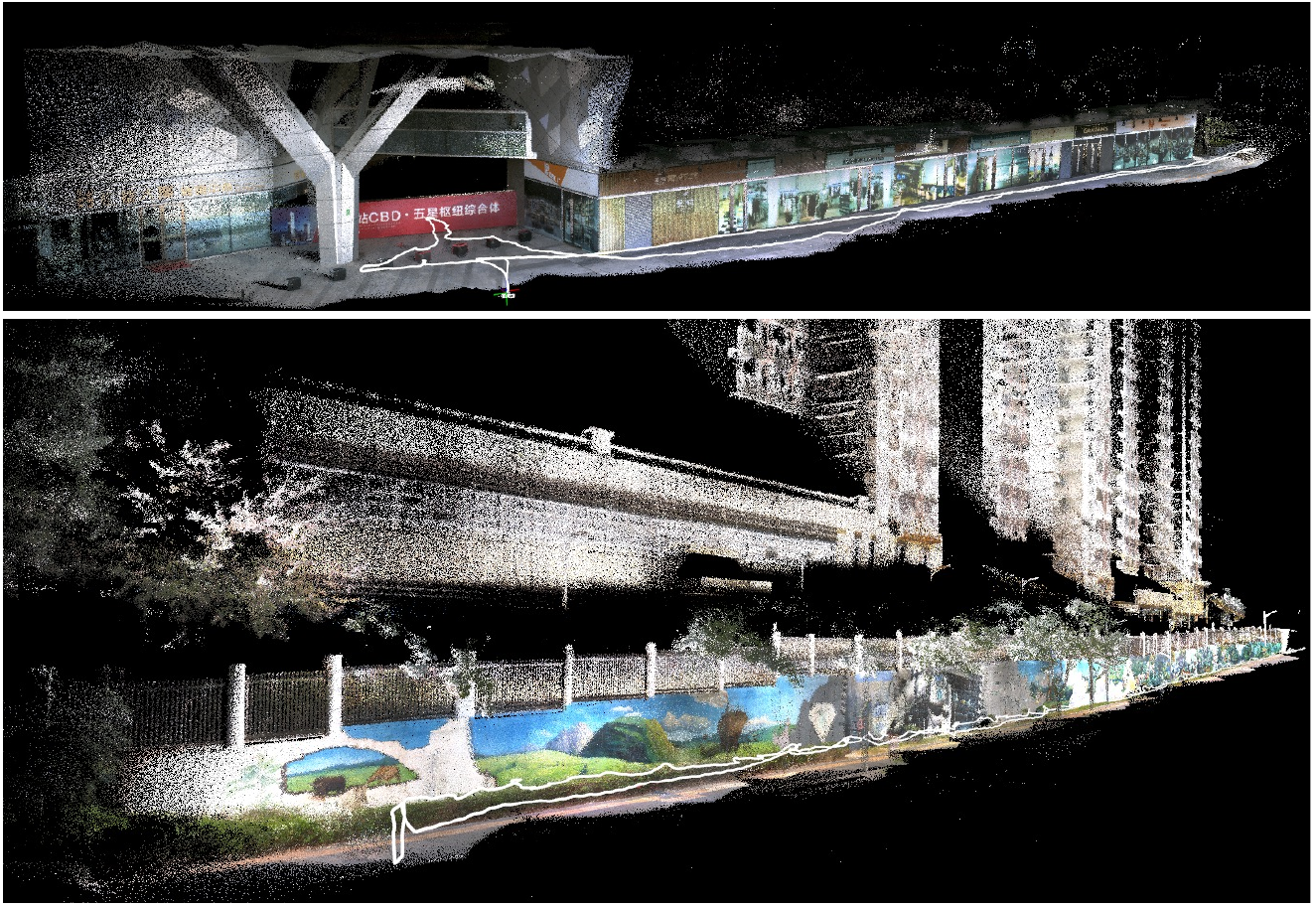}
    \caption{
    SLAM results of Gaussian-LIC2 (c2) in severely degenerated LiDAR scenarios of FAST-LIVO2. The odometry overcomes the degeneration and succeeds in returning to the origin. Note that as our system does not incorporate a detailed point cloud reconstruction module, the colored point cloud map is produced by downsampling and stitching the point clouds, leading to a somewhat coarse visual appearance.
    }
    \label{fig:exp_fastlivo2degenerate}
\end{figure}
\begin{figure}[t]
    \centering
    \includegraphics[width=\linewidth]{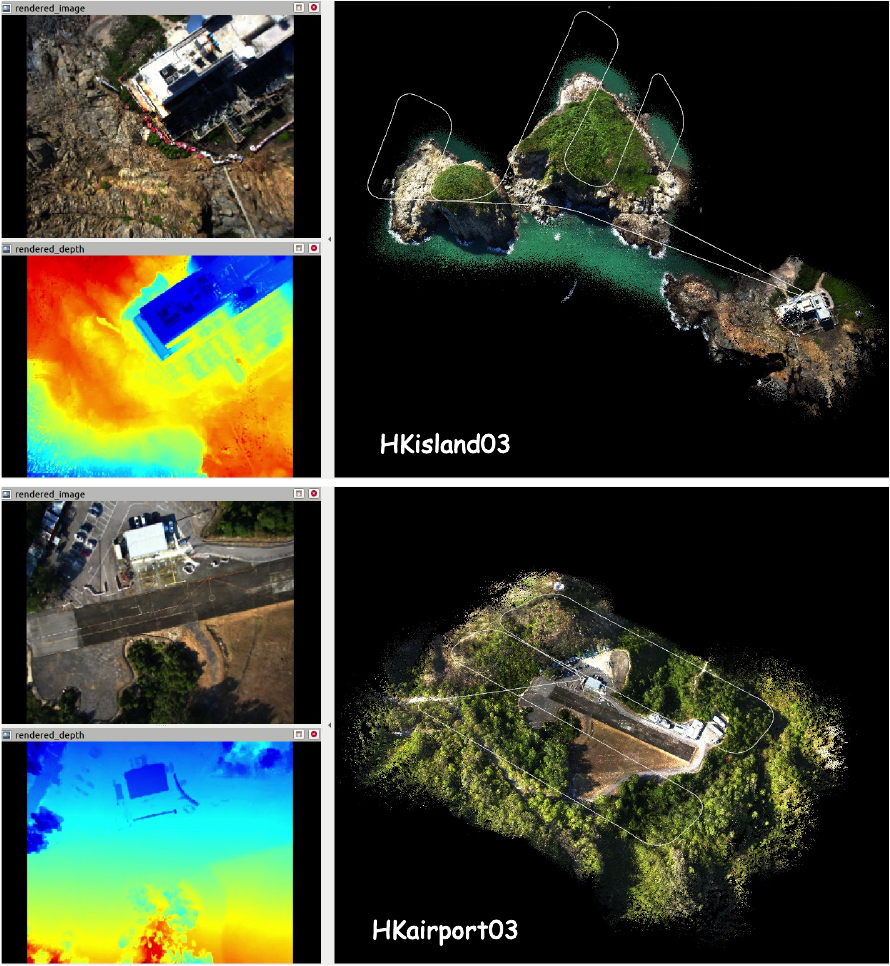}
    \caption{Real-time localization and photo-realistic mapping of Gaussian-LIC2 on the sequence \textit{HKisland03} and \textit{HKairport03} with a trajectory length of 1.95 km and 2.1 km, respectively.
    }
    \label{fig:island_airport}
\end{figure}
\begin{figure*}[t!]
    	\centering
    	\includegraphics[width=\textwidth]{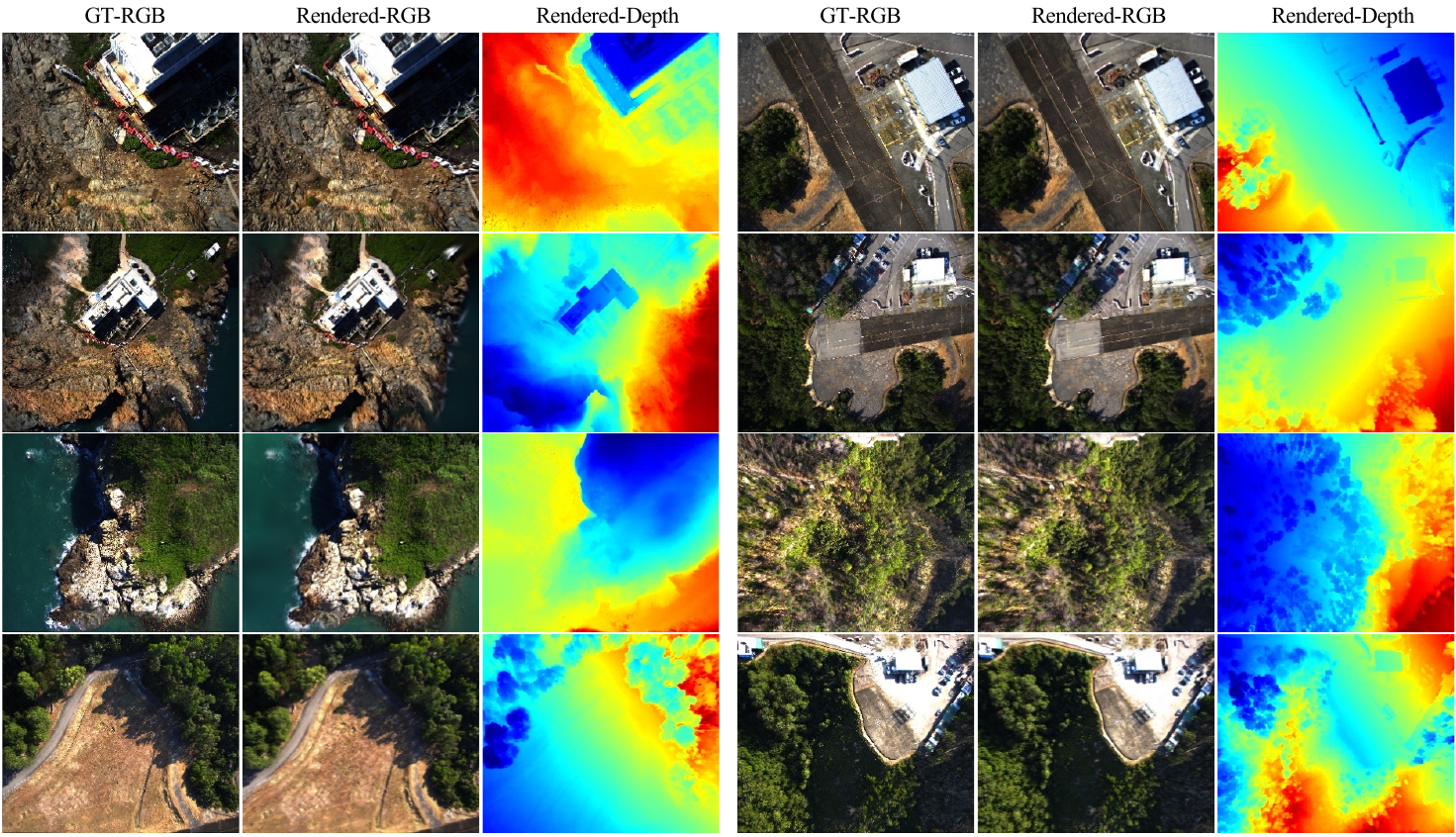}
    	\caption{Qualitative rendering results (in-sequence novel view) of Gaussian-LIC2 on the sequence \textit{HKisland03} and \textit{HKairport03}. 
     }
    \label{fig:mars_rendering}
\end{figure*}

\subsection{Comparison with Other Related Methods}
\label{sec:gslivm}
Tab.~\ref{tab:exp_gslivm} and Tab.~\ref{tab:exp_timecomparison} also present the comparison results with GS-LIVM~\cite{xie2024gs}, another recently released LiDAR-Inertial-Camera 3DGS-based SLAM
method.

\begin{table}[t]
\centering
\caption{PSNR results (in-sequence novel view) on our self-collected dataset. All methods use ground-truth poses.}
\label{tab:exp_gslivm}
\resizebox{\linewidth}{!}{
\begin{tabular}{@{}ccccc@{}}
\toprule
\multicolumn{1}{l}{} & \multicolumn{1}{l}{MM3DGS-SLAM} & \multicolumn{1}{l}{GS-LIVM} & \multicolumn{1}{l}{Gaussian-LIC} & \multicolumn{1}{l}{Gaussian-LIC2} \\ \midrule
Lecture\_Hall        & 19.54                           & 17.49                       & 21.60                            & \textbf{21.78}                    \\
Robot\_Center        & 18.83                           & 17.92                       & 22.89                            & \textbf{22.95}                    \\
Bell\_Tower\_01      & 18.13                           & 21.12                       & 25.38                            & \textbf{25.53}                    \\
Bell\_Tower\_02      & 16.49                           & 22.44                       & 26.42                            & \textbf{26.85}                    \\
Bell\_Tower\_03      & 16.61                           & 23.07                       & 27.29                            & \textbf{27.63}                    \\ \bottomrule
\end{tabular}
}
\end{table}
\begin{table}[t]
\centering
\caption{Runtime performance across different methods on the sequence
\textit{Bell\_Tower\_01} (span: 150 seconds). All approaches rely on
their estimated poses. Here is Gaussian-LIC2 (c1).}
\label{tab:exp_timecomparison}
\resizebox{\linewidth}{!}{
\begin{tabular}{@{}lcccc@{}}
\toprule
\multirow{2}{*}{}                & \multirow{2}{*}{PSNR$\uparrow$} & \multirow{2}{*}{Depth-L1 (m)$\downarrow$} & \multicolumn{2}{c}{Processing Time (s)$\downarrow$}                          \\ \cmidrule(l){4-5} 
                                 &                       &                               & \multicolumn{1}{l}{RTX 4090} & \multicolumn{1}{l}{RTX 3090} \\ \midrule
MonoGS                           & $\times$                     & $\times$                             & $\times$                            & $\times$                            \\
Co-SLAM                          & 13.26                 & 1.54                          & 1030                         & 1473                         \\
MM3DGS-SLAM                      & 16.07                 & 0.99                          & 2460                         & 3525                         \\
GS-LIVM                          & 21.05                 & 1.19                          & 103                          & 147                          \\
Gaussian-LIC                     & 25.36                 & 0.93                          & 102                          & 145                          \\
Gaussian-LIC2                    & \textbf{25.51}        & \textbf{0.27}                 & \textbf{90}                  & \textbf{128}                 \\
Gaussian-LIC2 w/o Tile Culling   & 25.48                 & 0.28                          & 109                          & 156                          \\
Gaussian-LIC2 w/o Per-Gaussian   & 25.51                 & 0.27                          & 145                          & 207                          \\
Gaussian-LIC2 w/o Sparse Adam    & 25.51                 & 0.27                          & 126                          & 180                          \\
Gaussian-LIC2 w/o Separated SH   & 25.51                 & 0.27                          & 98                           & 139                          \\
Gaussian-LIC2 w/o Efficient SSIM & 25.51                 & 0.27                          & 102                          & 146                          \\
Gaussian-LIC2 w/o WLPM           & 25.51                 & 0.27                          & 105                          & 152                          \\
Gaussian-LIC2 w/o CPU to GPU     & 25.51                 & 0.27                          & 100                           & 144                          \\
Gaussian-LIC2 w/o full           & 25.48                 & 0.28                          & 234                          & 335                          \\ \bottomrule
\end{tabular}
}
\end{table}
\begin{table}[t]
\centering
\caption{Runtime breakdown excluding mapping on the sequence
\textit{Bell\_Tower\_01} (span: 150 seconds). Here is Gaussian-LIC2 (c1). Values outside/inside the parentheses correspond to RTX 3090/RTX 4090, respectively. A scan contains LIC (LiDAR-Inertial-Camera) data spanning a duration of 0.1 s.}
\label{tab:exp_timecomparison2}
\resizebox{0.6\linewidth}{!}{
\begin{tabular}{@{}cc@{}}
\toprule
\textbf{Component}      & \textbf{Time per Scan (ms)} \\ \midrule
LIC Data Preprocessing  & 0.30                        \\
Trajectory Extension    & 0.01                        \\
Trajectory Optimization & 15.95                       \\
Visual Local Map Update & 2.07                        \\
Optical Flow Tracking   & 2.46                        \\
LiDAR Local Map Update  & 1.66                        \\
kNN Search              & 5.27                        \\
Marginalization         & 2.96                        \\
Render LiDAR Depth      & 1.29                        \\
Depth Completion        & 10.00 (23.06)              \\ \midrule
Total                   & 41.97 (55.03)                \\ \bottomrule
\end{tabular}
}
\end{table}

\subsection{Additional Results on Runtime}
\label{sec:runtime_ablation}

Tab.~\ref{tab:exp_timecomparison} displays the supplementary results of the runtime with different types of
NVIDIA GPUs. Note that except for tile culling, the other CUDA-related acceleration strategies hardly affect the mapping results. Tab.~\ref{tab:exp_timecomparison2} presents the runtime breakdown excluding mapping. For each 0.1 s LIC data segment, the tracking front-end consistently performs pose estimation within 0.1 s, achieving real-time performance.

\subsection{More Localization Results in Severely Degenerated Sequences from FAST-LIVO2}
\label{sec:degenerated_fastlivo2}

\begin{table}[t]
\centering
\caption{The start-to-end drift error (translation $|$ rotation) in severely degenerated LiDAR
scenarios of FAST-LIVO2.}
\label{tab:exp_fastlivo2degenerate}
\resizebox{\linewidth}{!}{
\begin{tabular}{
    l
    c@{\hspace{6pt}{$\;|\;$}\hspace{6pt}}c  
    c@{\hspace{6pt}{$\;|\;$}\hspace{6pt}}c  
    c@{\hspace{6pt}{$\;|\;$}\hspace{6pt}}c  
}
\toprule
& \multicolumn{2}{c}{CBD\_Building\_02} &
  \multicolumn{2}{c}{HIT\_Graffiti\_Wall\_01} &
  \multicolumn{2}{c}{Bright\_Screen\_Wall} \\
\midrule
Gaussian-LIC2-LIO  & 35.17m & 1.30° & 8.81m & 2.55° & 3.72m & 1.18° \\
FAST-LIVO2         & \textbf{0.01m}  & 0.77° & \textbf{0.03m} & \textbf{3.92°} & \textbf{0.02m} & 2.71° \\
Gaussian-LIC2 (c1) & 0.03m  & 0.88° & \textbf{0.03m} & 4.26° & 0.04m & 2.82° \\
Gaussian-LIC2 (c2) & \textbf{0.01m}  & \textbf{0.75°} & \textbf{0.03m} & 4.01° & \textbf{0.02m} & \textbf{2.54°} \\
\bottomrule
\end{tabular}
}
\end{table}

Tab.~\ref{tab:exp_fastlivo2degenerate} and Fig.~\ref{fig:exp_fastlivo2degenerate} show the supplementary results of Gaussian-LIC2 under severe LiDAR degeneration in sequences \textit{HIT\_Graffiti\_Wall}, \textit{Bright\_Screen\_Wall}, and \textit{CBD\_Building} from FAST-LIVO2~\cite{zheng2024fast}. Similar to the main text, Gaussian-LIC2 (c2) slightly outperforms Gaussian-LIC2 (c1) in sequences \textit{CBD\_Building\_02} and \textit{Bright\_Screen\_Wall} containing textureless walls.

\subsection{More Mapping Results in Large-Scale UAV Sequences from MARS-LVIG Dataset}
We additionally conduct experiments on the public MARS-LVIG~\cite{li2024mars} dataset, which provides data collected in large-scale natural environments using unmanned aerial vehicles. Fig.~\ref{fig:island_airport} presents visualization snapshots during the execution of Gaussian-LIC2, where ``rendered\_image'' and ``rendered\_depth'' denote the 3DGS renderings of the current frame.  Fig.~\ref{fig:mars_rendering} shows the in-sequence novel view synthesis results on the sequence \textit{HKisland03} and \textit{HKairport03}.

\begin{figure}[t]
    \centering
    \includegraphics[width=\linewidth]{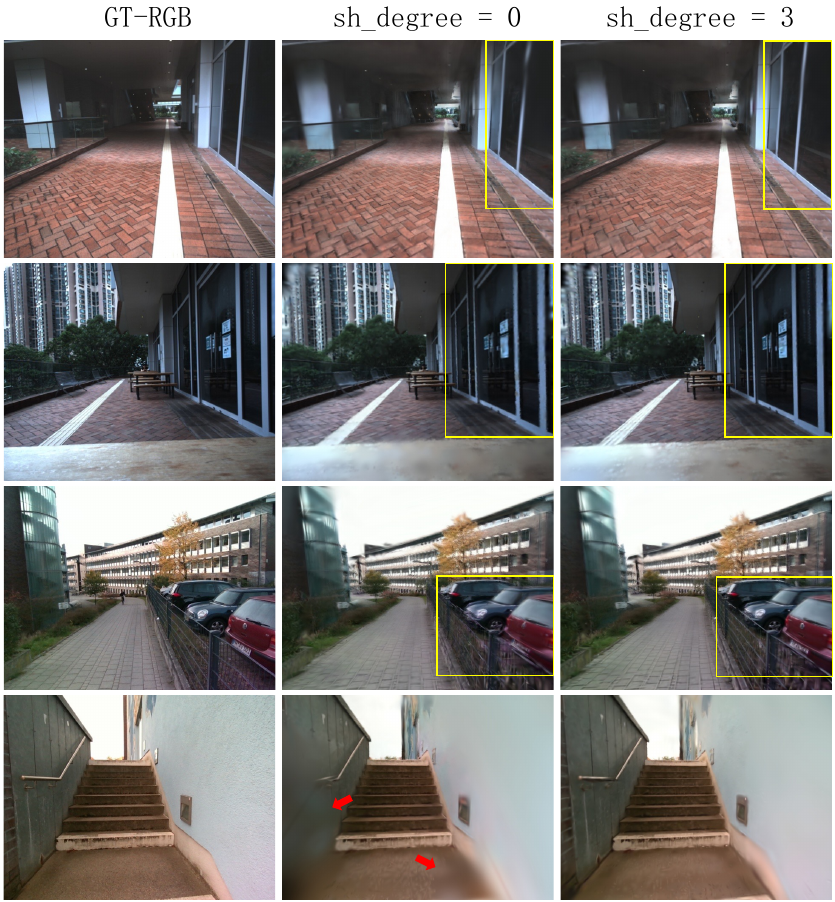}
    \caption{
    Qualitative rendering results of Gaussian-LIC2 with 0-order SH and 3rd-order SH. Regions with significant differences are highlighted by yellow boxes and red arrows.
    }
    \label{fig:exp_reflection}
\end{figure}

\begin{table}[htbp]
\centering
\caption{Ablation study on Adaptive Density Control (ADC).}
\label{tab:ablation_adc}
\resizebox{\linewidth}{!}{
\begin{tabular}{@{}lcccccc@{}}
\toprule
\multirow{2}{*}{Sequence} & 
\multicolumn{3}{c}{Gaussian-LIC2} & 
\multicolumn{3}{c}{Gaussian-LIC2 w/ ADC} \\
\cmidrule(lr){2-4} \cmidrule(l){5-7}
& {PSNR$\uparrow$} & {SSIM$\uparrow$} & {LPIPS$\downarrow$} & {PSNR$\uparrow$} & {SSIM$\uparrow$} & {LPIPS$\downarrow$} \\
\midrule
Retail\_Street    & \textbf{24.37} & \textbf{0.775} & \textbf{0.121} & 22.59 & 0.754 & 0.132 \\
CBD\_Building\_01 & \textbf{25.20} & \textbf{0.851} & \textbf{0.103} & 23.19 & 0.830 & 0.116 \\
\bottomrule
\end{tabular}
}
\end{table}
\begin{figure}[htbp]
    	\centering
    	\includegraphics[width=\linewidth]{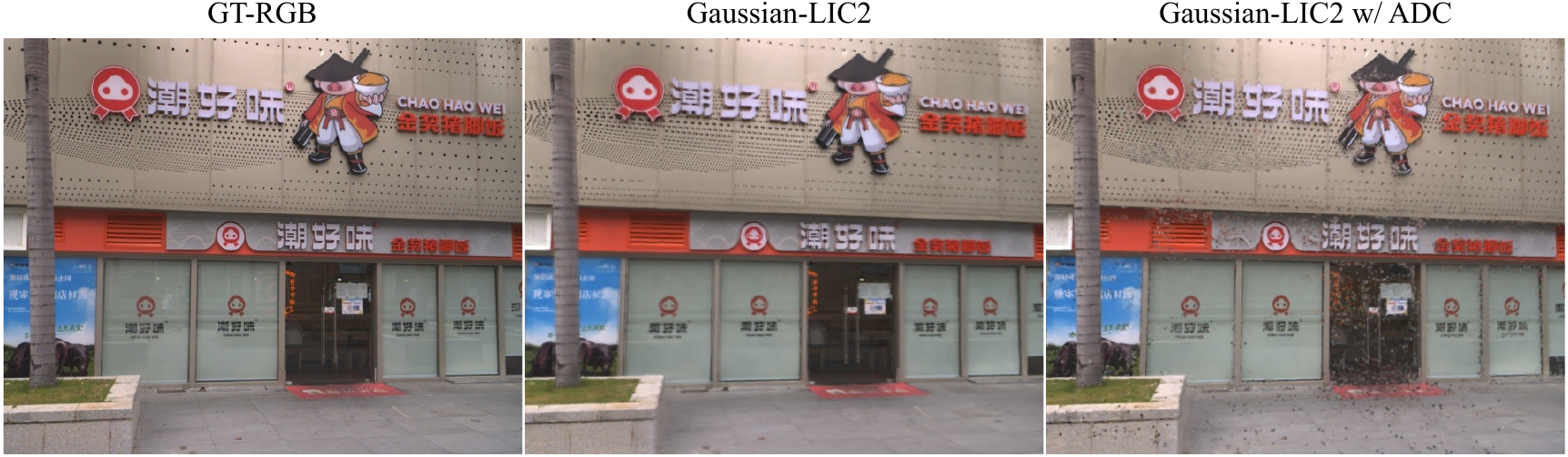}
    	\caption{
        Adaptive density control (ADC) struggles in incremental SLAM systems, often introducing floaters.
     }
    \label{fig:adc}
\end{figure}

\subsection{Ablation Study on Spherical Harmonics }
As shown in Fig.~\ref{fig:exp_reflection}, Gaussian-LIC2 with 3rd-order SH achieves better visual quality compared to the system that uses 0th-order SH, especially on non-Lambertian surfaces.

\subsection{Ablation Study on Adaptive Density Control}
 In incremental systems such as SLAM, Gaussians are added in successive batches, which results in different maturities among Gaussians from different batches. ADC tends to split and clone recently added Gaussians that have poor convergence, leading to many floaters and a reduction in visual quality, as depicted in Tab.~\ref{tab:ablation_adc} and Fig.~\ref{fig:adc}.

\end{document}